\documentclass[runningheads]{llncs}

\usepackage{subcaption}
\usepackage{amssymb}
\usepackage{graphicx}
\usepackage{enumerate}
\usepackage{algorithmic}
\usepackage{algorithm}
\usepackage{hyperref}
\usepackage{url}
\usepackage{amsmath}
\usepackage{dsfont}
\usepackage{pgfplots}
\usepackage{fca}
\usepackage{verbatim}

\urldef{\mailsa}\path|dignatov@hse.ru|
\urldef{\mailsb}\path|leonard.kwuida@bfh.ch|
\newcommand{\keywords}[1]{\par\addvspace\baselineskip
\noindent\keywordname\enspace\ignorespaces#1}

\usepackage{tikz}
\usepackage{tikz-qtree}

\usepackage{colortbl}
\usepackage{xcolor}

\definecolor{tomato}{RGB}{255,99,71}

\definecolor{lightseagreen}{RGB}{32,178,170}

\definecolor{darkkhaki}{RGB}{189,183,107}

\definecolor{Gray}{gray}{0.9}

\definecolor{obj-red}{rgb}{1,0,0}
\definecolor{obj-blue}{rgb}{0,0,1}
\definecolor{obj-green}{rgb}{0,0.5,0}
\definecolor{grey}{gray}{.5}

\newcommand{\bi}{\begin{itemize}}
\newcommand{\ei}{\end{itemize}}
\newcommand{\be}{\begin{enumerate}}
\newcommand{\ee}{\end{enumerate}}

\newcommand{\bd}{\begin{definition}}
\newcommand{\ed}{\end{definition}}

\newcommand{\K}{\context}

\makeatletter
\def\blfootnote{\gdef\@thefnmark{}\@footnotetext}
\makeatother

\begin{document}

\mainmatter  

\title{On Interpretability and Similarity in Concept-Based Machine Learning}

\titlerunning{On Interpretability and Similarity in Concept-Based Machine Learning}

%
%
\author{L\'eonard Kwuida\inst{1} and Dmitry I. Ignatov\inst{2,3}}
%

 \institute{Bern University of Applied Sciences, Bern, Switzerland\\
 \mailsb\\
 \and
 National Research University Higher School of Economics, Russian Federation\\
 \mailsa\\
 \and 
 St. Petersburg Department of Steklov Mathematical Institute of Russian Academy of Sciences, Russia}

%
%

\maketitle

\begin{abstract}
Machine Learning (ML) provides important techniques for classification and predictions. Most of these are black-box models for users and do not provide decision-makers with an explanation. For the sake of transparency or more validity of decisions, the need to develop explainable/interpretable ML-methods is gaining more and more importance. Certain questions need to be addressed: 

\begin{itemize}
\item How does an ML procedure derive the class for a particular entity? 
\item Why does a particular clustering emerge from a particular unsupervised ML procedure?
\item What can we do if the number of attributes is very large?
\item What are the possible reasons for the mistakes for concrete cases and models? 
\end{itemize}

For binary attributes, Formal Concept Analysis (FCA) offers techniques in terms of intents of formal concepts, and thus provides plausible reasons for model prediction. However, from the interpretable machine learning viewpoint, we still need to provide decision-makers with the importance of individual attributes to the classification of a particular object, which may facilitate explanations by experts in various domains with high-cost errors like medicine or finance. 

We discuss how notions from cooperative game theory can be used to assess the contribution of individual attributes in classification and clustering processes in concept-based machine learning. To address the 3rd question, we present some ideas on how to reduce the number of attributes using similarities in large contexts. 

\end{abstract}

\section{Introduction}

In the notes of this invited talk, we would like to give the reader a short introduction to Interpretable Machine Learning (IML) from the perspective of Formal Concept Analysis (FCA), which can be considered as a mathematical framework for concept learning, Frequent Itemset Mining (FIM) and Association Rule Mining (ARM). 

Among the variety of concept learning methods, we selected the rule-based JSM-method named after J.S.~Mill in its FCA formulation. Another possible candidate is Version Spaces. To stress the difference between concept learning paradigm and formal concept we used concept-based learning term in case of usage of FCA as a mathematical tool and language.

We assume, that interpretation by means of game-theoretic attribute ranking is also important in an unsupervised setting as well, and demonstrate its usage via attribution of stability indices of formal concepts (concept stability is also known as the robustness of closed itemset in the FIM community).

Being a convenient language for JSM-method (hypotheses learning) and Frequent Itemset Mining, its direct application to large datasets is possible only under a reasonable assumption on the number of attributes or data sparseness. Direct computation of the Shapley value for a given attribute also requires enumeration of almost all attribute subsets in the intent of a particular object or concept. One of the possibilities to cope with the data volume is approximate computations, while another one lies in the reduction of the number of attributes or their grouping by similarity.  

The paper is organised as follows. Section~\ref{sec:RelWork} observes several closely related studies and useful sources on FCA and its applications. Section~\ref{sec:SupLearn} is devoted to concept-based learning where formal intents are used as classification hypotheses and specially tailored Shapley value helps to figure out contributions of attributes in those hypotheses when a particular (e.g., unseen) object is examined. Section~\ref{sec:UnsupLearn} shows that the Shapley value approach can be used for attribution to stability (or robustness) of formal concepts, thus we are able to rank single attributes of formal intents (closed itemsets) in an unsupervised setting. Section~\ref{sec:AttrSimRed} sheds light on the prospects of usage attribute-based similarity of concepts and attribute reduction for possibly large datasets (formal contexts). Section~\ref{sec:Concl} concludes the paper.

\section{Related Work}\label{sec:RelWork}

Formal Concept Analysis is an applied branch of modern Lattice Theory suitable for knowledge representation and data analysis in various domains~\cite{Ganter:1999}. We refer the reader to a modern textbook on FCA with a focus on attribute exploration and knowledge extraction~\cite{Ganter:2016}, surveys on FCA models and techniques for knowledge processing and representation~\cite{Kuznetsov:2013,Poelmans:2013b} as well as on their applications~\cite{Poelmans:2013a}. Some of the examples in subsequent sections are also taken from a tutorial on FCA and its applications~\cite{Ignatov:2014a}.

Since we deal with interpretable machine learning, we first need to establish basic machine learning terminology in FCA terms. In the basic case, our data are Boolean object-attribute matrices or formal contexts, which are not necessarily labeled w.r.t. a certain target attribute. Objects can be grouped into clusters (concept extents) by their common attributes, while attributes compose a cluster (concept intent) if they belong to a certain subset of objects. The pairs of subsets of objects and attributes form the so-called formal concepts, i.e. maximal submatrices (w.r.t. of rows and attribute permutations) of an input context full of ones in its Boolean representation. Those concepts form hierarchies or concept lattices (Galois lattices), which provide convenient means of visualisation and navigation and enables usage of suprema and infima for incomparable concepts.

The connection between well-known concept learning techniques (for example, Version Spaces, and decision tree induction) from machine learning and FCA was well established in~\cite{Ganter:2003,Kuznetsov:2004}. Thus Version Spaces studied by T.~Mitchell~\cite{Mitchell:1977} also provides hierarchical means for hypotheses learning and elimination, where hypotheses are also represented as conjunctions of attributes describing the target concept. Moreover, concept lattices can be used for searching for globally optimal decision trees in the domains where we should not care about the trade-off between time spent for the training phase and reached accuracy (e.g., medical diagnostics) but should rather focus on all valid paths in the global search space~\cite{Belohlavek:2009,Kashnitsky:2016}.

In case we deal with unsupervised learning, concept lattices can be considered as a variant of hierarchical clustering where one has the advantage to use multiple inheritance in both bottom-up and top-down directions~\cite{Carpineto:96,Valtchev:2000,Stumme:2001,Bocharov:2016}. Another fruitful property of formal concepts allows one not only to receive a cluster of objects without any clue why they are similar but to reveal objects' similarity in terms of their common attributes. This property allows considering a formal concept as bicluster~\cite{Mirkin:1996,Ignatov:2012,Kaytoue:2014}, i.e. a biset of two clusters of objects and attributes, respectively.

Another connection between FCA and Frequent Itemset Mining is known for years~\cite{Pasquier:1999,Lakhal:2005}. In the latter discipline, transactions of attributes are mined to find items frequently bought together~\cite{Agrawal:93}. The so-called closed itemsets are used to cope with a huge number of frequent itemsets for large input transaction bases (or contexts), and their definition coincides with the definition of concept intents (under the choice of constraint on the concept extent size or itemset support). Moreover, attribute dependencies in the form of implications and partial implications~\cite{Luxenburger:1991} are known as association rules, which appeared later in data mining as well~\cite{Agrawal:93}\footnote{One of the earlier precursors of association rules can be also found in~\cite{Hajek:1966} under the name of ``almost true implications''}. 

This is not a coincidence that we discuss data mining, while stressed interpretability and machine learning in the title.  Historically, data mining was formulated as a step of the Knowledge Discovery in Databases process that is ``the nontrivial process of identifying valid, novel, potentially useful, and ultimately understandable patterns in data.''~\cite{Fayyad:96}. While understandable patterns are a must for data mining, in machine learning and AI in general, this property should be instantiated as something extra, which is demanded by analysts to ease decision making as the adjectives explainable (AI) and interpretable (ML) suggest~\cite{Molnar:2019}. 

To have a first but quite comprehensive reading on interpretable ML we suggest a freely available book~\cite{Molnar:2019}, where the author states that ``Interpretable Machine Learning refers to methods and models that make the behaviour and predictions of machine learning systems understandable to humans''.

The definition of interpretability may vary from the degree to which a human can understand the cause of a decision to the degree to which a human can consistently predict the model's result.

The taxonomy of IML methods has several aspects. For example, models can be roughly divided into \emph{intrinsic} and \emph{post hoc} ones. The former include simpler models like short rules or sparse linear models, while among the latter black-box techniques with post hoc processing after their training can be found. Some researchers consistently show that in case of the necessity to have interpretable models, one should not use post hoc techniques for black-box models but trust naturally interpretable models~\cite{Rudin:2019}. Another aspect is the universality of the method, the two extremes are \emph{model-specific} (the method is valid for only one type of models) and or \emph{model-agnostic} (all models can be interpreted with the method). There is one more important aspect, whether the method is suitable for the explanation of the model's predictions for a concrete object (\emph{local method}) or it provides an interpretable picture for the entire model (\emph{global method}). Recent views on state-of-the-art techniques and practices can be found in~\cite{Caruana:2020,Kaur:2020}.

FCA provides interpretable patterns a priori since it deals with such understandable patterns as sets of attributes to describe both classes (by means of classification rules or implications) and clusters (e.g., concept intents). However, FCA theory does not suggest the (numeric) importance of separate attributes. Here, a popular approach based on Shapley value from Cooperative Game Theory~\cite{Shapley:1953} recently adopted by the IML community may help~\cite{Strumbelj:2014,Lundberg:2017,Kadyrov:2019}.

The main idea of Shapley value based approaches in ML for ranking separate attributes is based on the following consideration: each attribute is considered as a player in a specific game-related to classification or regression problem and attributes are able to form (winning) coalitions. The importance of such a player (attribute) is computed over all possible coalitions by a combinatorial formula taking into account the number of winning coalitions where without this attribute the winning state is not reachable.  

One of the recent popular implementations is SHAP library~\cite{Lundberg:2017}, which however cannot be directly applied to our concept-based learning cases: JSM-hypotheses and stability indices.  The former technique assumes that unseen objects can be left without classification or classified contradictory when for an examined object there is no hypothesis for any class or there are at least two hypotheses from different classes~\cite{Finn:1983,Kuznetsov:1990}. This might be an especially important property for such domains as medicine and finance where wrong decisions may lead to regrettable outcomes. We can figure out what are the attributes of the contradictory hypotheses we have but which attributes have the largest positive or negative impact on the classification is still unclear without external means. The latter case of stability indices, which were originally proposed for ranking JSM-hypotheses by their robustness to the deletion of object subsets from the input contexts (similarly to cross-validation)~\cite{Kuznetsov:1991,Kuznetsov:2007a}, is considered in an unsupervised setting. Here, supervised interpretable techniques like SHAP are not directly applicable. To fill the gap we formulated two corresponding games with specific valuation functions used in the Shapley value computations. 

Mapping of the two proposed approaches onto the taxonomy of IML methods says that in the case of JSM-hypotheses it is an intrinsic model, but applying Shapley values on top of it is post hoc. At the same time, this concrete variant is rather model-specific since it requires customisation. This one is local since it explains the classification of a single object. As for attribution of concept stability, this one is definitely post hoc,  model-specific, and if each pattern (concept) is considered separately this one is rather local but since the whole set of stable concepts can be attributed it might be considered as a global one as well.

It is important to note that one of the stability indices was rediscovered in the Data Mining community and known under the name of the robustness of closed itemsets~\cite{Tatti:2011,Kuznetsov:2018} (where each transaction/object is kept with probability $\alpha=0.5$). So, the proposed approach also allows attribution of closed itemsets.

Classification and direct computation of Shapley values afterwards might be unfeasible for large sets of attributes~\cite{Caruana:2020}. So, we may think of approximate ways to compute Shapley values~\cite{Strumbelj:2014} or pay attention to attribute selection, clarification, and reduction known in the FCA community. We would like to draw the reader's attention to scale coarsening as feature selection tools~\cite{Ganter:2008} and a comparative survey on FCA-based attribute reduction techniques~\cite{Konecny:2017,Konecny:2018}. However, we prefer to concentrate on attribute aggregation by similarity~\footnote{Similarity between concepts is discussed in~\cite{Eklund:2012}} as an attribute reduction technique which will not allow us to leave out semantically meaningful attributes even if they are highly-correlated and redundant in terms of extra complexity paid for their processing otherwise. 

The last note on related works, which is unavoidable when we talk about IML, is the relation to Deep Learning (DL) where black-box models predominate~\cite{Shrikumar:2017}. According to the textbook~\cite{Goodfellow:2016}, ``Deep Learning is a form of machine learning that enables computer to learn from experience and understand the world in terms of a hierarchy of concepts.'' The authors also admit that there is no need for a human computer operator to formally specify all the knowledge that the computer needs and obtained hierarchy of concepts allows the computer to learn complicated concepts by building them out of simpler ones. The price of making those concepts intelligible for the computer but not necessary for a human is paid by specially devised IML techniques in addition to DL models.  

Since FCA operates with concept hierarchies and is extensively used in human-centric applications~\cite{Poelmans:2013a}, the question ``What can FCA do for DL?'' is open. For example, in~\cite{Rudolph:2007} closure operators on finite sets of attributes were encoded by a three-layered feed-forward neural network, while in~\cite{Kuznetsov:2017} the authors were performing neural architecture search based on concept lattices to avoid overfitting and increase the model interpretability. 

\section{Supervised Learning: From Hypotheses to Attribute Importance}\label{sec:SupLearn}

In this section, we discuss how interpretable concept-based learning for JSM-method can be achieved with Shapley Values following our previous study on the problem~\cite{Ignatov:2020a}.
Let us start with a piece of history of inductive reasoning. In XIX century, John Stuart Mill proposed several schemes of inductive reasoning. Let us consider, for example, the Method of Agreement~\cite{Mill:1843}:
``If two or more instances of the phenomenon under investigation have only one circumstance in common, ... [it] is the cause (or effect) of the given phenomenon.''    

The JSM-method (after J.S. Mill) of hypotheses generation proposed by Viktor K. Finn in the late 1970s is an attempt to describe induction in purely deductive form~\cite{Finn:1983}. 
This new formulation was introduced in terms of many-valued many-sorted extension of the First Order Predicate Logic~\cite{Kuznetsov:2005}. 

This formal logical treatment allowed usage of the JSM-method as a machine learning technique~\cite{Kuznetsov:1991}. While further algebraic redefinitions of the logical predicates to express similarity of objects as an algebraic operation allowed the formulation of JSM-method as a classification technique in terms of formal concepts~\cite{Kuznetsov:1996,Kuznetsov:2005}.

\subsection{JSM-hypotheses in FCA}

In FCA, a formal concept consists of an extent and an intent. The intent is formed by all attributes that describe the concept, and the extent contains all objects belonging to the concept. In  FCA,  the  JSM-method is known as rule-based learning from positive and negative examples with rules in the form ``concept intent $\to$ class''.

Let a formal context $\mathbb{K} :=(G, M, I)$ be our universe, where the binary relation $I \subseteq G \times M$ describes if an object $g\in G$ has an attribute $m\in M$. For $A\subseteq G$ and $B\subseteq M$ the derivation (or Galois) operators are defined by:
		\[ A' =\{\,m\in M \mid \forall a\in A\, aIm\,\} \text{ and }  B' =\{\,g\in G \mid \forall b\in B\, gIb\,\} . \]
A (formal) concept is a pair $(A,B)$  with  $A\subseteq G$, $B\subseteq M$ such that $A'=B$ and $B'=A$.  We call $B$ its intent and $A$ its extent. An implication of the form $H\to m$ holds if all objects having the attributes in $H$ also have the attribute $m$, i.e. $H' \subseteq m'$.

The set of all concepts of a given context $\context$ is denoted by $\BGMI$; the concepts are ordered by the ``to be a more general concept'' relation as follows: $(A,B) \geq (C,D) \iff C \subseteq A \ (\mbox{equivalently } B \subseteq D)$. 

The set of all formal concepts $\BGMI$ together with the introduced relation form the \emph{concept lattice}, which line diagram is useful for visual representation and navigation through the concept sets.  

Let  $w\notin M$ be a \emph{target attribute}, then $w$ partitions $G$ into three subsets:
\bi
\item \emph{positive examples}: $G_+\subseteq G$ of objects known to satisfy $w$,
\item \emph{negative examples}: $G_-\subseteq G$ of objects known not to have $w$,
\item \emph{undetermined examples}: $G_{\tau}\subseteq G$  of objects for which it
remains unknown whether they have the target attribute or do not have it.
\ei
This partition gives rise to three subcontexts $\K_\varepsilon:=(G_\varepsilon,M,I_\varepsilon)$ with $ \varepsilon \in \{-, +, \tau\}$. 


\bi

\item The \emph{positive context} $\K_+$ and the \emph{negative context} $\K_-$ form the training set called by \emph{learning context}:
$$\K_\pm = (G_+ \cup G_-, M \cup \{w\}, I_+ \cup I_- \cup G_+\times \{w\}).$$

\item The subcontext $\K_\tau$ is called the \emph{undetermined context} and is used to predict the class of not yet classified objects. 

\ei

The whole \emph{classification context} is the context
$$\K_c = (G_+ \cup G_- \cup G_\tau, M \cup \{w\}, I_+ \cup I_- \cup I_\tau  \cup G_+\times \{w\}).$$ The derivation operators in the subcontexts $\K_\varepsilon$ are denoted by  $(\cdot)^+$ $(\cdot)^-$, and $(\cdot)^\tau$, respectively. The goal is to classify the objects in $G_{\tau}$ with respect to $w$. 

To do so let us form the positive and negative hypotheses as follows. 
A \emph{positive hypothesis} $H\subseteq M$ ($H\neq \emptyset$)
is a intent of $\K_+$ that is not contained in the intent of a
negative example; i.e. $H^{++} = H$  and $H'\subseteq G_+\cup G_{\tau}$ ($H \to w$).
A \emph{negative hypothesis} $H\subseteq M$ ($H\neq \emptyset$)
is an intent of $\K_-$ that is not contained in the intent of a
positive example; i.e. $H^{--} = H$  and $H'\subseteq G_-\cup G_{\tau}$  ($H \to \overline{w}$).

An intent of $\K_+$ that is contained in the intent of a negative example is called a  \emph{falsified (+)-generalisation}. A \emph{falsified (-)-generalisation\emph} is defined in a similar way.

To illustrate these notions we use the credit scoring context in Table~\ref{tbl:scor}~\cite{Ignatov:2014b}. Note that we use \emph{nominal scaling} to transform many-valued context to one-valued context~\cite{Ganter:1999} with the following attributes, $Ma$, $F$ (for two genders), $Y$, $MI$, $O$ (for young, middle, and old values of the two-valued attribute Age , resp.), $HE$, $Sp$, $SE$ (for higher, special, and secondary education, resp.), $Hi$, $L$, $A$ (for high, low, and average salary, resp.), and $w$ and $\overline{w}$ for the two-valued attribute Target.

\begin{table}[htbp]
	\centering
	\caption{Many-valued classification context for credit scoring}\label{tbl:scor}
		\begin{tabular}{|c||c|c|c|c|c|}
		\hline
		G / M & Gender & Age & Education & Salary & Target\\
		\hline
		\hline
1 & Ma & young & higher & high & $+$\\
2 & F & middle & special & high & $+$\\
3 & F & middle & higher & average & $+$\\
4 & Ma & old & higher & high & $+$\\
		\hline
5 & Ma & young & higher & low & $-$\\
6 & F & middle & secondary & average &$-$\\
7 & F & old & special & average &$-$\\
8 & Ma  & old &  secondary & low &$-$\\
		\hline
9 & F & young & special & high & $\tau$\\
10 & F & old & higher & average & $\tau$\\
11 & Ma & middle & secondary & low & $\tau$\\
12 & Ma & old & secondary & high & $\tau$\\
		\hline
		\end{tabular}
\end{table}
%
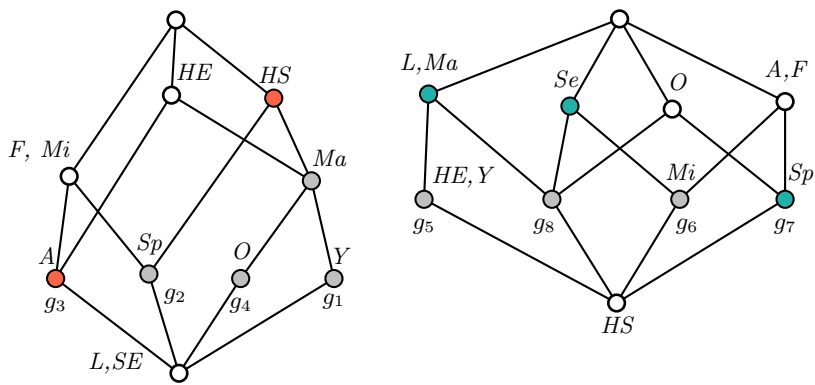
\begin{figure}[h]
\begin{minipage}[h]{0.45\linewidth}  
  \centering
	\begin{picture}(150,160)
\unitlength 0.20mm
\begin{diagram}{270}{260}
\Node{1}{155}{252}
\Node{2}{157}{17}
\Node{3}{84}{148}
\Node{4}{75}{80}
\Node{5}{137}{83}
\Node{6}{152}{202}
\Node{7}{245}{145}
\Node{8}{260.0}{80.0}
\Node{9}{198}{80.0}
\Node{10}{220.0}{200.0}
\Edge{6}{1}
\Edge{7}{10}
\Edge{3}{1}
\Edge{4}{3}
\Edge{5}{10}
\Edge{2}{4}
\Edge{4}{6}
\Edge{2}{9}
\Edge{8}{7}
\Edge{9}{7}
\Edge{2}{5}
\Edge{5}{3}
\Edge{10}{1}
\Edge{7}{6}
\Edge{2}{8}
\NoDots
\centerObjbox{4}{0}{13}{$g_3$}
\rightObjbox{5}{10}{10}{$g_2$}
\centerObjbox{8}{0}{10}{$g_1$}
\centerObjbox{9}{0}{13}{$g_4$}
\leftAttbox{2}{25}{0}{L,SE}
\leftAttbox{3}{0}{10}{F, Mi}
\leftAttbox{4}{0}{10}{A}
\centerAttbox{5}{0}{13}{Sp}
\rightAttbox{6}{3}{10}{HE}
\rightAttbox{7}{0}{10}{Ma}
\centerAttbox{8}{3}{10}{Y}
\centerAttbox{9}{0}{13}{O}
\centerAttbox{10}{0}{10}{HS}
\CircleSize{11}
\end{diagram}
\put(-195,80){\color{tomato}{\circle*{10}}}%
\put(-50,200){\color{tomato}{\circle*{10}}}%
\put(-133,83){\color{lightgray}{\circle*{10}}}%
\put(-72,80){\color{lightgray}{\circle*{10}}}%
\put(-10,80){\color{lightgray}{\circle*{10}}}%
\put(-25,145){\color{lightgray}{\circle*{10}}}%
\end{picture}
 \end{minipage} 
\hspace{2mm}
\begin{minipage}[h]{0.45\linewidth}
\begin{picture}(100,100)
  \unitlength 0.20mm
\begin{diagram}{210}{200}
\Node{1}{160.0}{200}
\Node{2}{158.0}{11.0}
\Node{3}{33.0}{150.0}
\Node{4}{115.0}{80.0}
\Node{5}{30.0}{80.0}
\Node{6}{270.0}{145.0}
\Node{7}{200.0}{80.0}
\Node{8}{270.0}{80.0}
\Node{9}{127.0}{142.0}
\Node{10}{195.0}{140.0}
\Edge{2}{5}
\Edge{4}{10}
\Edge{7}{6}
\Edge{7}{9}
\Edge{2}{7}
\Edge{8}{10}
\Edge{9}{1}
\Edge{5}{3}
\Edge{6}{1}
\Edge{3}{1}
\Edge{10}{1}
\Edge{2}{4}
\Edge{8}{6}
\Edge{2}{8}
\Edge{4}{3}
\Edge{4}{9}
\NoDots
\centerObjbox{4}{-5}{13}{$g_8$}
\centerObjbox{5}{0}{13}{$g_5$}
\centerObjbox{7}{5}{13}{$g_6$}
\centerObjbox{8}{0}{13}{$g_7$}
\centerAttbox{2}{0}{-20}{HS}
\centerAttbox{3}{0}{13}{L,Ma}
\centerAttbox{5}{25}{8}{HE,Y}
\centerAttbox{6}{0}{13}{A,F}
\centerAttbox{7}{0}{13}{Mi}
\centerAttbox{8}{10}{10}{Sp}
\centerAttbox{9}{-3}{13}{Se}
\centerAttbox{10}{3}{13}{O}
\CircleSize{11}
\end{diagram}
\put(-177,150){\color{lightseagreen}{\circle*{10}}}%
\put(-83,142){\color{lightseagreen}{\circle*{10}}}%
\put(60,80){\color{lightseagreen}{\circle*{10}}}%
\put(-10,80){\color{lightgray}{\circle*{10}}}%
\put(-95,80){\color{lightgray}{\circle*{10}}}%
\put(-180,80){\color{lightgray}{\circle*{10}}}%
\end{picture}
  \end{minipage} 
 \caption{The line diagrams of the lattice of positive context (left) and the lattice of negative context (right).}
  \label{fig:posneglatts}
\end{figure}
For example, the intent of the red node labelled by the attribute $A$ in the left line diagram (Fig.~\ref{fig:posneglatts}), is $\{A, Mi, F, HE \}$, and this is not contained in the intent of any node labelled by the objects $g_5, g_6, g_7$, and $g_8$. So we believe in the rule $H\to w$. Note that the colours of the nodes in Fig.~\ref{fig:posneglatts} represent different types of hypotheses: the red ones correspond to minimal hypotheses (cf. the definition below), the see green nodes correspond to negative hypotheses, while light grey nodes correspond to non-minimal positive and negative hypotheses for the left and the right line diagrams, respectively.

The undetermined examples $g_\tau$ from $G_{\tau}$ are classified according to the following rules: 
\bi
\item If $g^{\tau}_\tau$ contains a positive, but no negative hypothesis, then $g_\tau$ is \emph{classified positively}. 

\item If $g^{\tau}_\tau$ contains a negative, but no positive hypothesis, then $g_\tau$ is \emph{classified negatively}. 

\item If $g^{\tau}_\tau$ contains both negative and positive hypotheses, or if $g^{\tau}_\tau$ 
 does not contain any hypothesis, then this object classification is \emph{contradictory} or \emph{undetermined}, respectively. 
\ei

To perform classification by the aforementioned rules, it is enough to have only \emph{minimal hypotheses} (w.r.t. $\subseteq$) of both signs.

Let $\mathcal{H}_+$ (resp.  $\mathcal{H}_-$) be the set of minimal positive (resp. minimal negative) hypotheses. Then,
\begin{small}
\begin{align*}
    \mathcal{H}_+ = \big\{ \{F, Mi, HE, A\}, \{HS\} \big\} \text{ and }
    \mathcal{H}_- = \big\{ \{F, O, Sp, A\}, \{Se\}, \{Ma, L\} \big\} .
\end{align*}
\end{small}

We proceed to classify the four undetermined objects below.

\bi
\item $g_9'=\{F, Y, Sp, HS\}$ contains the positive hypothesis $\{HS\}$, and no negative hypothesis. Thus, $g_9$ is classified positively. 
\item  $g_{10}'=\{F, O, HE, A\}$ does not contain neither positive nor negative hypotheses. Hence, $g_{10}$ remains undetermined. 
\item $g_{11}'=\{Ma, Mi, Se, L\}$ contains two negative hypotheses: $\{Se\}$ and $\{Ma, L\}$, and no positive hypothesis. Therefore, $g_{11}$ is classified negatively. 
\item $g_{12}'=\{Ma, O, Se, HS\}$ contains the negative hypothesis $\{Se\}$ and the positive hypothesis $\{HS\}$, which implies that $g_{12}$ remains undetermined.
\ei

Even though we have a clear explanation of why a certain object belongs to one of the classes in terms of contained positive and negative hypotheses, the following question arises: Do all attributes play the same role in the classification of certain examples? If the answer is no, then one more question appears: How can we rank attributes with respect to their importance in classifying examples, for example, $g_{11}$ with attributes $ Ma, Mi, Se$, and $L$? Game Theory offers several indices for such comparison: e.g., the Shapley value and the Banzhaf index. For the present contribution, we concentrate on the use of Shapley values. 










\subsection{Shapley values and JSM-hypotheses}

To answer the question ``What are the most important attributes for classification of a particular object?'' in our case, we follow to basic recipe studied in \cite{Strumbelj:2014,Lundberg:2017,Molnar:2019}.

To compute the Shapley value for an example $x$ and an attribute $m$, one needs to define $f_x(S)$, the expected value of the model prediction conditioned on a subset $S$ of the input attributes.

\begin{equation}\label{eq:shapley}
\phi_m = \sum_{S \subseteq M \setminus \{m\}} \frac{|S|!(|M| - |S| -1)!}{|M|!} \left ( f_x(S \cup \{m\}) - f_x(S) \right ),
\end{equation}
where $M$ is the set of all input attributes and $S$ a certain coalition of players, i.e. set of attributes. 

Let $\K_c = (G_+ \cup G_- \cup G_\tau, M \cup \{w\}, I_+ \cup I_- \cup I_\tau  \cup G_+\times \{w\})$ be our classification context, and $\mathcal{H_+}$ (resp. $\mathcal{H_-}$) the set of minimal positive (resp. negative) hypotheses of $\K_c$. 

Since we deal with hypotheses (i.e. sets of attributes) rather than compute the expected value of the model's prediction, we can define a valuation function $v$ directly. For $g \in G$, the \emph{Shapley value of an attribute} $m \in g'$:
\begin{equation}\label{eq:shapley_hyp}
\varphi_m (g) = \sum_{S \subseteq g' \setminus \{m\}} \frac{|S|!(|g'| - |S| -1)!}{|g'|!} \left ( v(S \cup \{m\}) - v(S) \right ), 
\end{equation}
where
$$v(S)=\begin{cases}
    1,  &  \exists H_+ \in \mathcal{H_+}: H_+ \subseteq S \mbox{ and } \forall H_- \in \mathcal{H_-}: H_- \not\subseteq S, \\
    -1,  &  \exists H_- \in \mathcal{H_-}: H_- \subseteq S  \mbox{ and } \forall H_+ \in \mathcal{H_+}: H_+ \not\subseteq S \\
    0,  &  \text{otherwise}
 \end{cases} $$

The Shapley value $\varphi_{m}(g)$ is set to 0 for every $m \in M\setminus g'$. The Shapley vector for a given object $g$ is denoted by $\varPhi(g)$. To differentiate between the value in cases when $m \in M\setminus g'$ and $m \in g'$, we will use decimal separator as follows, 0 and 0.0, respectively.

For the credit scoring context, the minimal positive and the negative hypotheses are
\begin{align*}
    \mathcal{H}_+ &= \{\{F, Mi, HE, A\}, \{HS\}\}; \
    \mathcal{H}_- = \{\{F, O, Sp, A\}, \{Se\}, \{M, L\}\}.
\end{align*}

The Shapley values for JSM-hypotheses have been computed with our freely available Python scripts\footnote{\url{https://github.com/dimachine/Shap4JSM}} for the objects in $G_\tau$:

\begin{itemize}
\item $g_9'=\{F, Y, Sp, HS\} \supseteq \{HS\}$, and $g_9$ is classified positively. $\varphi_\text{HS}(g_9)=1$ and 
 and its Shapley vector is
$\varPhi(g_9)=(0, 0.0, 0.0, 0, 0, 0, 0.0, 0, 1.0, 0, 0) \mbox{ .}$ 
%
\item $g_{10}'=\{F, O, HE, A\}$ and $g_{10}$ remains undetermined. Its Shapley vector is $\varPhi(g_{10})=(0, 0.0, 0, 0, 0.0, 0.0, 0, 0, 0, 0.0, 0) \mbox{ .}$ 
\item $g_{11}'=\{Ma, Mi, Se, L\}\supseteq \{Se\},\ \{Ma, L\}$. Its Shapley vector is \\ 
${\varPhi(g_{11})=(-1/6, 0, 0, 0.0, 0, 0, 0, -2/3, 0, 0, -1/6)}  \mbox{ .}$ 
\item $g_{12}'=\{Ma, O, Se, HS\}\supseteq \{HS\},\ \{Se\}$. $\varphi_\text{Se}(g_{12})=-1$, $\varphi_\text{HS}(g_{12})=1$. Its Shapley vector is  $\varPhi(g_{12})=(0.0, 0, 0, 0, 0.0, 0, 0, -1.0, 1.0, 0, 0) \mbox{ .}$
\end{itemize}

Let us examine example $g_{11}$. Its attribute  $Mi$ has zero importance according to the Shapley value approach since it is not in any contained hypothesis used for the negative classification. The most important attribute is $Se$, which is alone two times more important than the attributes $Ma$ and $L$ together. It is so, since the attribute $Se$, which is the single attribute of the negative hypothesis $\{Se\}$, forms more winning coalitions $S \cup\{Se\}$ with $v(S \cup \{Se\}) - v(S)=1$ than  $Ma$ and $L$, i.e. six vs. two. Thus, $\{Se\}\uparrow \setminus \{Ma, L\} \uparrow = \{\{Se\}, \{Ma, Se\}, \{Mi, Se\}, \{Se, L\},  \{Mi, Se, L\},$ $\{Ma, Mi, Se\}\} $\footnote{$S\uparrow$  is the up-set of $S$ in the Boolean lattice $(\mathcal{P}\{Ma, Mi, Se, L\},\subseteq)$} are such winning coalitions for $Se$, while $\{Ma, L\}$, $\{Ma, Mi, L\}$,  are those for $Ma$ and $L$. 

The following properties hold:
\begin{theorem}[\cite{Ignatov:2020a}]
The Shapley value, $\varphi_m(g)$, of an attribute $m$ for the JSM-classification of an object $g$,  fulfils the following properties:
\begin{enumerate}
      \item $\sum\limits_{m \in g'}\varphi_m(g)=1$ if $g$ is classified positively;
      \item $\sum\limits_{m \in g'}\varphi_m(g)=-1$ if $g$ is classified negatively.
    \item $\sum\limits_{m \in g'}\varphi_m(g)=0$ if $g$ is classified contradictory or undetermined.
 \end{enumerate}
\end{theorem}

The last theorem expresses the so-called \emph{efficiency property} or axiom~\cite{Shapley:1953}, where it is stated that the sum of Shapley values of all players in a game is equal to the total pay-off of their coalition, i.e. $v(g')$ in our case. 

It is easy to check $\varphi_m(g)=0$ for every $m \in g'$ that does not belong to at least one positive or negative hypothesis contained in $g'$. Moreover, in this case for any $S \subseteq g'\setminus \{m\}$ it also follows $v(S) = v(S \cup \{m\})$ and these attributes are called \emph{null} or \emph{dummy players}~\cite{Shapley:1953}.

We also performed experiments on the Zoo dataset\footnote{\url{https://archive.ics.uci.edu/ml/datasets/zoo}}, which includes 101 examples (animals) and their 17 attributes along with the target attribute (7 classes of animals).
The attributes are binary except for the number of legs, which can be scaled nominally and treated as categorical. 
  
We consider a binary classification problem where \emph{birds} is our positive class, while all the rest form the negative class.

There are  19 positive examples (birds) and  80 negative examples since we left out two examples for our testing set, namely, chicken and warm. 
The hypotheses are  
$\mathcal{H}_+ = \big\{\{ feathers, eggs, backbone, breathes, legs_2, tail  \}\big\}$ and 
\begin{small}
\begin{align*}
\mathcal{H}_-= &\big\{\{venomous\},
\{eggs, aquatic, predator, legs_5\},
\{legs_0\},
\{eggs, legs_6\},\\
& \{predator, legs_8\},
\{hair, breathes \},
\{milk, backbone, breathes \},
\{legs_4 \},\\
& \{toothed, backbone\}
\big\}.
 \end{align*}
\end{small}

The intent $aardvark'=\{hair, milk, predator, toothed, backbone, breathes, legs_4, catsize\}$ contains four negative hypotheses and no positive one.

\footnotesize
The Shapley vector for the \textbf{aardvark} example is
$$(-0.1,
 0,
 0,
 -0.0167,
 0,
 0,
 0.0,
 -0.1,
 -0.133,
 -0.133,
 0,
 0,
 -0.517,
 0,
 0,
 0,
 0,
 0,
 0,
 0,
 0.0) \mbox{ .}$$

\textbf{Backbone}, \textbf{breathes}, and \textbf{four legs} are the most important attributes with 
values -0.517, -0.133, and -0.133, respectively, while \textbf{catsize} is not important in terms of Shapley value.

\begin{figure}[ht!] 
   \centering
    \includegraphics[width=0.6\linewidth]{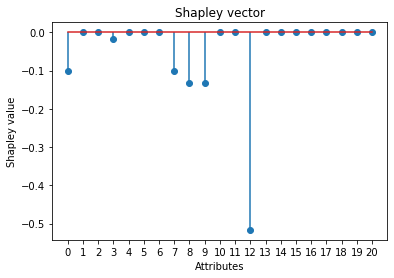} 
  \caption{The Shapley vector diagram for the \textbf{aardvark} example} \label{fig:lollipop} 
\end{figure}

A useful interpretation of classification results could be an explanation for true positive or true negative cases.  However, in the case of our test set both examples, \textbf{chicken} and \textbf{warm}, are classified correctly as bird and non-bird, respectively. Let us have a look at their Shapley vectors. Our test objects have the following intents: 
$$chicken'=\{feathers, eggs, airborne, backbone, breathes, legs_2, tail, domestic\}$$
and
$$warm'=\{eggs, breathes, legs_0\} .$$

\begin{figure}[ht!] 
   \centering
   \begin{minipage}[h]{0.49\linewidth}  
  \centering
    \includegraphics[width=1\linewidth]{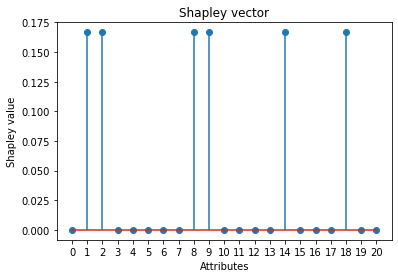}
    \end{minipage}
     \begin{minipage}[h]{0.49\linewidth}  
  \centering
    \includegraphics[width=1\linewidth]{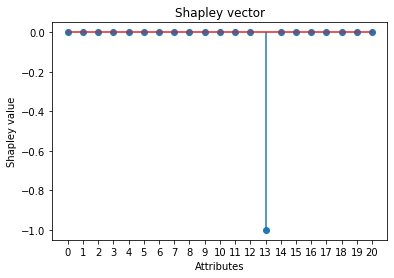} 
      \end{minipage}
  \caption{The Shapley vector diagram for the \textbf{chicken} (left) and \textbf{warm} (right) examples} \label{fig:lollipop} 
\end{figure}

Thus, for the \textbf{chicken} example all six attributes that belong to the single positive hypothesis have equal Shapley values, i.e. 1/6. The attributes \textbf{airborne} and \textbf{domestic} have zero importance. The \textbf{warm} example has only one attribute with non-zero importance, i.e. the absence of legs with importance -1. It is so since the only negative hypothesis, $\{legs_0\}$, is contained in the object intent. 



  



\section{Unsupervised Learning: Contribution to Stability and Robustness}\label{sec:UnsupLearn}

(Intensional) stability indices were introduced to rank the concepts (intents) by their robustness under objects deletion and provide evidence of the non-random nature of concepts~\cite{Roth:2006}. The extensional stability index is defined as the proportion of intent subsets generating this intent; it shows the robustness of the concept extent under attributes deletion~\cite{Roth:2006}. Our goal here is to find out whether all attributes play the same role in the stability indices. To measure the importance of an attribute for a concept intent, we compare generators with this attribute to those without it. In this section, we demonstrate how Shapley values can be used to assess attribute importance for concept stability.


\subsection{Stability indices of a concept}
Let $\mathbb{K} :=(G, M, I)$ be a formal context. For any closed subset $X$ of attributes or objects, we denote by $\mathrm{gen}(X)$ the set of generating subsets of $X$.  The \emph{extensional stability index}~\cite{Roth:2006} of a concept $(A,B)$ is 
	\begin{align*}
	\sigma_e(A,B) &:
	= \frac{|\{Y \subseteq B \mid Y''=B\}|}{2^{|B|}} = \frac{|\mathrm{gen}(B)|}{2^{|B|}}.
	\end{align*} 
We can also restrict to generating subsets of equal size. The extensional stability index of the $k$-th level of $(A,B)$ is
	$$J_k(A,B):=|\{Y \subseteq B\mid |Y|=k, Y''=B\}| \Big/\binom{|B|}{k}.$$

\subsection{Shapley vectors of intents for concept stability}
Let $(A,B)$ be a concept of $(G,M,I)$ and $m\in B$. We define an indicator function by 
$$v(Y)= 1 \mbox{ if } Y''=B \text{ and }  \ Y\neq \emptyset, \mbox{ and } v(Y) = 0 \text{ otherwise}. $$ 
Using the indicator $v$, the Shapley value of $m \in B$ for the stability index of the concept $(A,B)$ is defined by:

\begin{equation}\label{eq:shapley_stab}
	\varphi_m(A,B):=\frac{1}{|B|}\sum\limits_{Y\subseteq B\setminus \{m\}} \frac{1}{\binom{|B|-1}{|Y|}}\Big(v(Y \cup \{m\})-v(Y)\Big) .
\end{equation}

  The Shapley vector of $(A,B)$ is then $\left(\varphi_m(A,B)\right)_{m\in B}$.
  An equivalent formulation is given using upper sets of minimal generators~\cite{Ignatov:2020b}. In fact, for $m \in X_m \in \mathrm{mingen}(B)$ and $m \notin X_{\overline{m}}\in \mathrm{mingen}(B)$, we have  
	\begin{equation*}\label{eqn:shap-stab_0}
	\varphi_m(A,B)=\frac{1}{|B|}\sum\limits_{D\sqcup\{m\} \in \bigcup X_m\uparrow\setminus \bigcup X_{\overline{m}}\uparrow} \frac{1}{\binom{|B|-1}{|D|}},
	\end{equation*}
where $\sqcup$ denotes the disjoint union, $X_m$ and $X_{\overline{m}}$ the minimal generators of $B$ with and without $m$, respectively. 
	
\begin{figure}[htbp]
\centering
	\includegraphics[width=.25\linewidth]{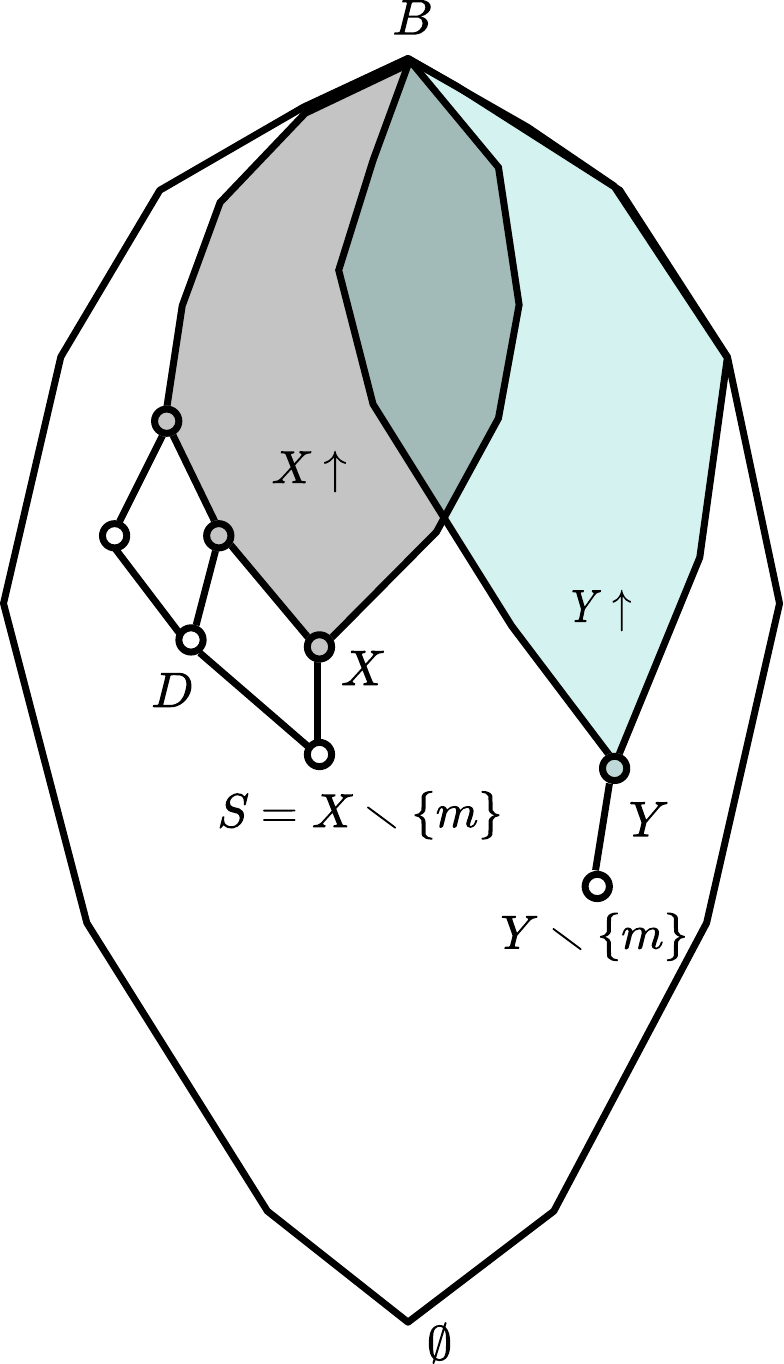} \hspace*{2cm}
	\includegraphics[width=.25\linewidth]{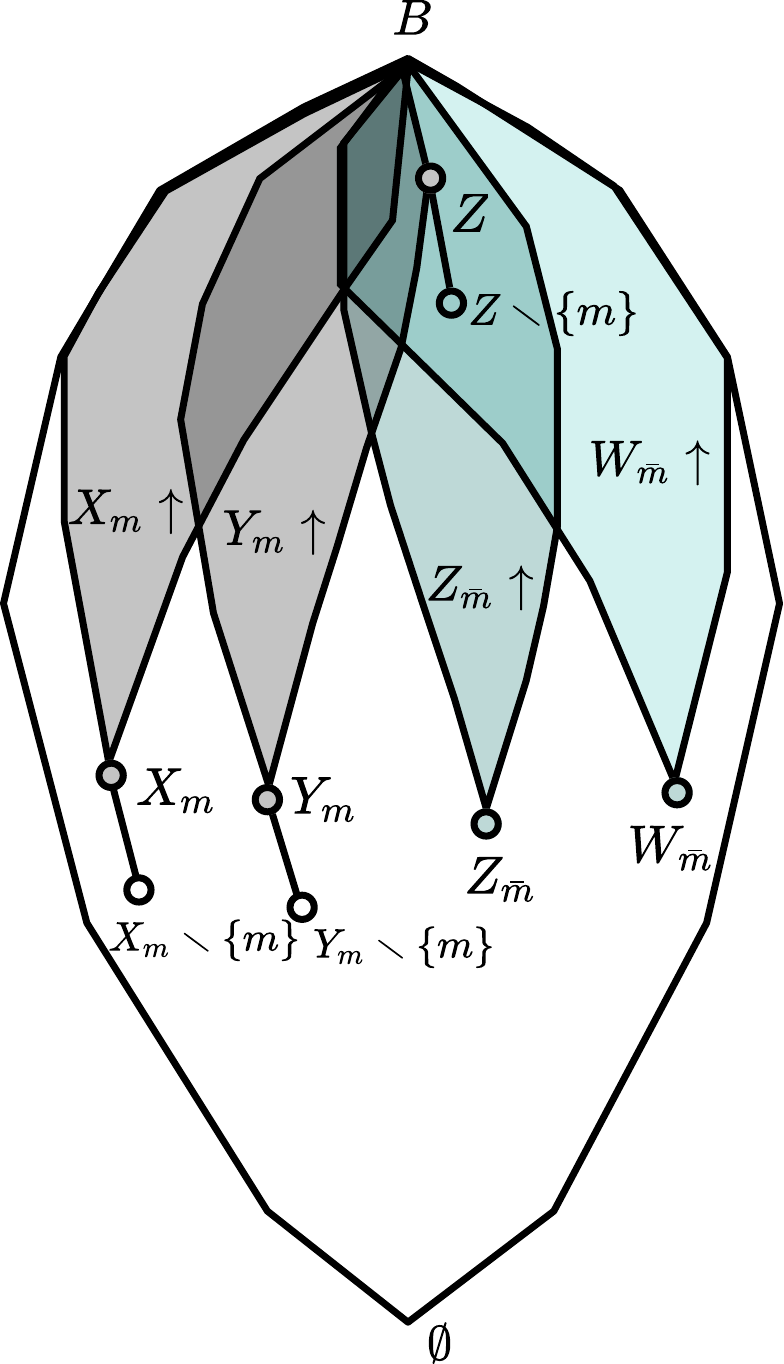}
\caption{Computing Shapley vectors for concept stability}
\end{figure}

To compute $\varphi_m$, additional simplifications are useful: 
\begin{theorem}[\cite{Ignatov:2020b}] 
Let $(A,B)$ be a concept and $m \in B$.
\begin{itemize}
    \item[(i)] $\varphi_m(A,B)=\sum\limits_{k=1}^{|B|} \frac{J_k(A,B)}{k} - \sum\limits_{D \subseteq B\setminus\{m\}}\frac{1}{|D|\binom{|B|-1}{|D|}}v(D)$.
    \item[(ii)] If $m \in X_m \in \mathrm{mingen}(B)$ and $Y \subseteq B\setminus \{m\}$ with $(A,B)\prec (Y',Y)$ then
    \begin{equation*} 
	\varphi_m(A,B)=\frac{1}{|B|}\sum\limits_{D \in \bigcup [X_m\setminus{\{m\}},Y]} \frac{1}{\binom{|B|-1}{|D|}}.
	\end{equation*}
	\item[(iii)] If $m \in X \in \mathrm{mingen}(B)$ and $|\mathrm{mingen}(B)|=1$, then
	\begin{equation}\label{eqn:shap-stab-mingen}
	\varphi_m(A,B)=\sum\limits_{k=1}^{|B|} \frac{J_k(A,B)}{k}=\frac{1}{|X|} \ .
	\end{equation}
\end{itemize}
\end{theorem}

To illustrate the importance of attributes in concept stability, we consider the 
the fruits context~\cite{Kuznetsov:2004}, where we extract the subcontext with the first four objects (Table~\ref{tab:fruits}). 

\begin{table}[]
\caption{A many-valued context of fruits}
\label{tab:fruits}
\begin{center}
\begin{tabular}{|c|c|cccc|}

\hline

& G $\setminus$ M & color & f{i}rm & smooth & form \\
\hline
1& apple & yellow  & no& yes & round \\
2& grapefruit & yellow & no& no  & round \\
3& kiwi & green  & no& no& oval\\
4& plum & blue  & no& yes& oval \\

\hline
\end{tabular}
\end{center}

\end{table}

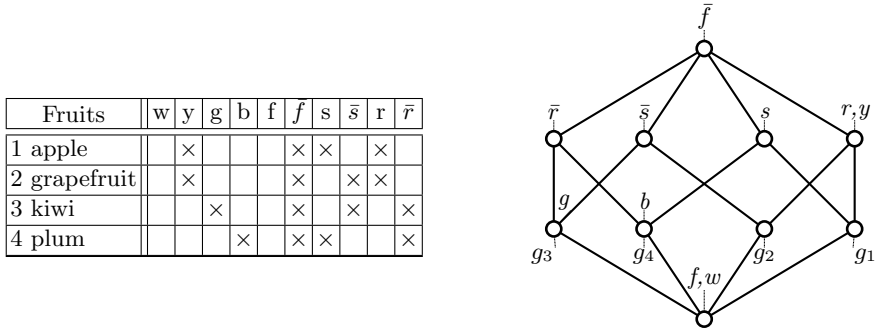
\begin{figure}
    \centering
\begin{minipage}{0.49\linewidth}
\begin{cxt}%
\cxtName{Fruits}%
\att{w}
\att{y}
\att{g}
\att{b}
\att{f}
\att{$\bar{f}$}
\att{s}
\att{$\bar{s}$}
\att{r}
\att{$\bar{r}$}
\obj{.x...xx.x.}{1 apple}
\obj{.x...x.xx.}{2 grapefruit}
\obj{..x..x.x.x}{3 kiwi}
\obj{...x.xx..x}{4 plum}
\end{cxt}
\end{minipage}
\begin{minipage}{0.49\linewidth}
\begin{picture}(100,130)
\unitlength 0.20mm
\begin{diagram}{270}{200}
\Node{1}{160.0}{200}
\Node{2}{160.0}{20}
\Node{3}{60.0}{140}
\Node{4}{60.0}{80}
\Node{5}{120.0}{80}
\Node{6}{260.0}{140}
\Node{7}{200.0}{80}
\Node{8}{260.0}{80}
\Node{9}{120.0}{140}
\Node{10}{200.0}{140}
\Edge{9}{1}
\Edge{4}{9}
\Edge{10}{1}
\Edge{4}{3}
\Edge{2}{5}
\Edge{8}{6}
\Edge{3}{1}
\Edge{7}{6}
\Edge{2}{4}
\Edge{5}{3}
\Edge{5}{10}
\Edge{2}{8}
\Edge{7}{9}
\Edge{8}{10}
\Edge{6}{1}
\Edge{2}{7}
\leftObjbox{4}{0}{13}{$g_3$}
\centerObjbox{5}{0}{13}{$g_4$}
\centerObjbox{7}{0}{13}{$g_2$}
\rightObjbox{8}{0}{13}{$g_1$}
\centerAttbox{1}{0}{13}{$\bar{f}$}
\centerAttbox{2}{0}{20}{f,w}
\centerAttbox{3}{0}{13}{$\bar{r}$}
\leftAttbox{4}{-10}{13}{g}
\centerAttbox{5}{0}{13}{b}
\centerAttbox{6}{0}{13}{r,y}
\centerAttbox{9}{0}{13}{$\bar{s}$}
\centerAttbox{10}{0}{13}{s}
\CircleSize{10}
\end{diagram}
\end{picture}
\end{minipage}
    \caption{A scaled fruits context and the line diagram of its concept lattice}
    \label{fig:fruits_latt}
\end{figure}

After scaling we get the binary context and its concept lattice diagram (Fig.~\ref{fig:fruits_latt}). 

\noindent
For each concept, the stability index $\sigma_e$ and its Shapley vector $\phi$ are computed. 

\begin{table}[]
    \caption{The concepts of fruits context and their stability indices along with Shapley vectors}
    \label{tab:my_label}
\centering
\setlength{\tabcolsep}{5pt}
\renewcommand{\arraystretch}{1.2}
\begin{tabular}{|c|l|c|c|}
\hline
Concepts & $\sigma_e$ & $\Phi$   
\\ \hline
 $(\{4\}, \{b, \bar{f}, s, \bar{r}\})$ &  0.625 & (2/3, 0.0, 1/6, 1/6)   
 \\ \hline

$(\{3\}, \{g, \bar{f}, \bar{s}, \bar{r}\})$ & 0.625 & 
(2/3, 0.0, 1/6, 1/6) 
\\ \hline

$(\{3, 4\}, \{\bar{f}, \bar{r}\})$ & 0.5 & (0.0, 1.0) 
\\ \hline

$(\{2\}, \{y, \bar{f}, \bar{s}, r\})$ &  0.375 & (1/6, 0.0, 2/3, 1/6) 
\\ \hline

$(\{2, 3\}, \{\bar{f}, \bar{s} \})$ & 0.5 & (0.0, 1.0) 
\\ \hline

$(\{1\}, \{y, \bar{f}, s, r\})$ & 0.375 & (1/6, 0.0, 2/3, 1/6) 
\\ \hline

$(\{1, 4\}, \{\bar{f}, s\})$ & 0.5 & (0.0, 1.0) 
\\ \hline

$(\{1, 2\}, \{y, \bar{f}, r\})$ & 0.75 & (0.5, 0.0, 0.5) 
\\ \hline

$(\{1, 2, 3, 4\}, \{\bar{f}\})$ &  1 & (0.0) 
\\ \hline
\end{tabular}

 \medskip

\begin{tabular}{|c|}
\hline
$\sigma_e(\emptyset, \{w, y, g, b, f, \bar{f}, s, \bar{s}, r, \bar{r}\})=0.955$\\
\hline
$\Phi=(0.256, 0.069, 0.093, 0.093, 0.260, 0.0, 0.052, 0.052, 0.069, 0.052)$ \\
\hline
 \end{tabular}
\end{table}


For the Zoo dataset we obtain 357 concepts in total. The top-3 most stable are $c_1, c_2, c_3$ with extent stability indices: 
  $\sigma_e(G,\emptyset)=1$, $\sigma_e(\emptyset,M)=0.997$, 
$\sigma_e(A, A')=0.625$, respectively,  where \\
$A' = \{feathers, eggs, backbone, breathes, legs_2, tail\}$ and 

\noindent $A=\{11, 16,  20,  21,  23, 33, 37, 41, 43, 56, 57, 58, 59, 71, 78, 79, 83, 87, 95, 100\} \ .$ 

\begin{footnotesize}
\begin{figure}[ht!] 
\begin{minipage}[b]{0.5\linewidth}
    \includegraphics[width=1\linewidth]{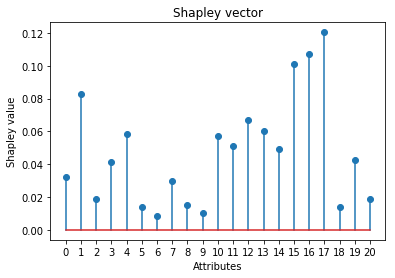} 
  \end{minipage}
  \begin{minipage}[b]{0.5\linewidth}
     \includegraphics[width=1\linewidth]{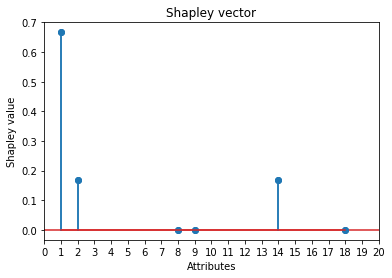} 
  \end{minipage}
 \caption{The Shapley vector for concept $c_2=(\emptyset,M)$ (left) and $c_3$ (right)} 
\end{figure}
The most important attributes are \textbf{six legs}, \textbf{eight legs}, \textbf{five legs}, \textbf{feathers}, and \textbf{four legs} for $c_2$, and 
\textbf{feathers},  \textbf{eggs}, and  \textbf{two legs} for $c_3$, w.r.t. to the Shapley vectors.
\end{footnotesize}

The demo is available on GitHub\footnote{\url{https://github.com/dimachine/ShapStab/}}.
Shapley values provide a tool for assessing the attribute importance of stable concepts. Comparison with other (not only Game-theoretic) techniques for local interpretability is desirable. We believe that the attribute importance can be lifted at the context level, via an aggregation, and by then offer a possibility for attribute reduction, similar to the principal component analysis (PCA) method.










\section{Attribute Similarity and Reduction}\label{sec:AttrSimRed}
Computation of attribute importance could lead to ranking the attributes of the context, and by then classifying the attributes with respect to their global importance, similar to principal component analysis. Therefore cutting off at a certain threshold could lead to attribute reduction in the context. Other methods leading to attributes reduction are based on their granularity, an ontology or an is-a taxonomy, by using coarser attributes. Less coarse attributes are then put together by going up in the taxonomy and are considered to be similar. In the present section, we briefly discuss the effect of putting attributes together on the resulting concept lattice. Doing this leads to the reduction of the number of attributes, but not always in the reduction of the number of concepts. 

Before considering such compound attributes, we would like to draw the readers' attention to types of data weeding that often overlooked outside of the FCA community~\cite{Priss:2011,Konecny:2017,Konecny:2018}, namely, clarification and reduction. 

\def\neu#1{{\bf #1}}
\def\KK{\mathbb{K}}
\def\NN{\mathbb{N}}
\def\RR{\mathbb{R}}
\def\J{\mathop{\mbox{\rm J}}}
\def\I{\mathop{\mbox{\rm I}}}
\def\cal#1{\mathcal{#1}}
\def\frak#1{\mathfrak{#1}}
\def\Ext{\mathop{\mbox{\rm Ext}}}

\subsection{Clarification and reduction}

A context $(G, M, I)$ is called \emph{clarified}~\cite{Ganter:1999}, if for any objects $g,h \in G$ from $g'=h'$ it always follows that $g=h$ and, similarly, $m'=n'$ implies $m=n$ for all $m,n \in M$. A clarification consists in removing duplicated lines and columns from the context. 
This context manipulation does not alter the structure of the concept lattice, though objects with the same intents and attributes with the same extents are merged, respectively.

The structure of the concept lattice remains unchanged in case of removal of \emph{reducible attributes} and \emph{reducible objects}~\cite{Ganter:1999}; An attribute $m$ is reducible if it is a combination of other attributes, i.e. $m'=Y'$ for some $Y \subseteq M$ with $m \not \in Y$. Similarly, an object $g$ is reducible if $g'=X'$ for some $X \subseteq G$ with $g \not \in X$.
For example, full rows ($g'=M$) and full columns ($m'=G$) are always reducible.

However, if our aim is a subsequent interpretation of patterns, we may wish to keep attributes (e.g. in aggregated form), rather than leaving them out before knowing their importance.

\subsection{Generalised attributes}\label{sec:gen attributes}
As we know, FCA is used for conceptual clustering and helps discover patterns in terms of clusters and rules. However, the number of patterns can explode with the size of an input context. Since the main goal is to maintain a friendly overview of the discovered patterns, several approaches have been investigated to reduce the number of attributes without loss of much information~\cite{Priss:2011,Konecny:2017}.  One of these suggestions consists in using is-a taxonomies. Given a taxonomy on attributes, how can we use it to discover \emph{generalised} patterns in the form of clusters and rules?  If there is no taxonomy, can we (interactively) design one?  We will discuss different scenarios of grouping attributes or objects, and the need of designing similarity measures for these purposes in the FCA setting. 

\begin{figure}[ht]
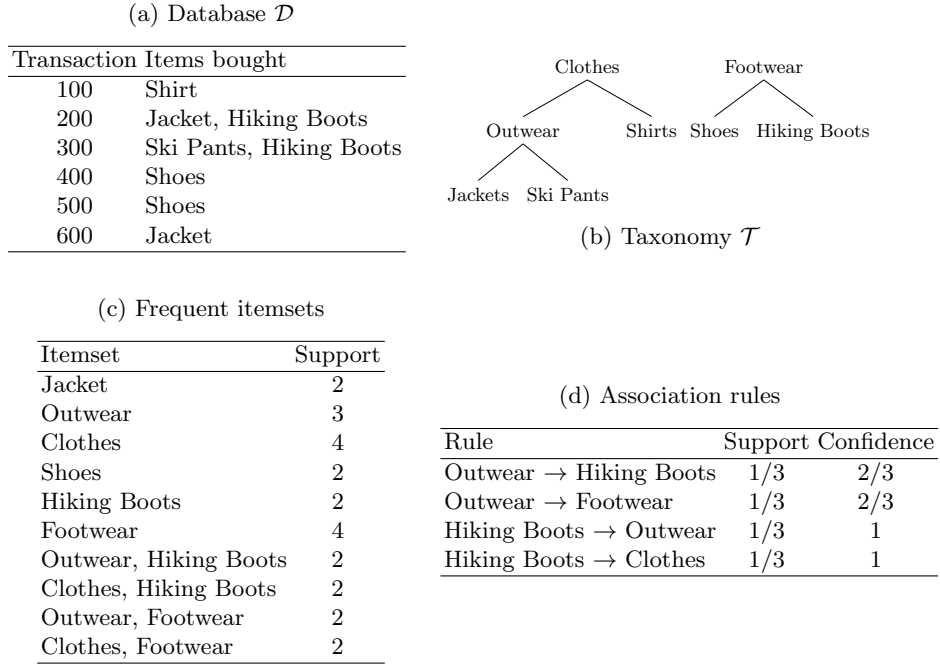
 
  \begin{subfigure}[b]{0.5\linewidth}
    \centering
    \caption{Database $\mathcal D$} 
    \label{fig7:a}
    \begin{tabular}{cl}
    \hline
         Transaction & Items bought \\
    \hline     
         100 & Shirt \\
         200 & Jacket, Hiking Boots \\
         300 & Ski Pants, Hiking Boots \\
         400 & Shoes \\
         500 & Shoes \\
         600 & Jacket \\
    \hline 
    \end{tabular}
    \vspace{4ex}
  \end{subfigure}
  \begin{subfigure}[b]{0.5\linewidth}
\begin{minipage}[h]{0.5\linewidth} 
\resizebox{1.9\linewidth}{!}{%
\Tree [.Clothes [.Outwear [.Jackets ] [.{Ski Pants} ] ]
[.Shirts ]]
\Tree [.Footwear [.Shoes ] [.{Hiking Boots} ]]
}\hfill
\end{minipage}
    \caption{Taxonomy $\mathcal{T}$} 
    \label{fig7:b} 
    \vspace{4ex}
  \end{subfigure} 
  \begin{subfigure}[b]{0.5\linewidth}
    \centering
    \caption{Frequent itemsets} 
    \label{fig7:c} 
     \begin{tabular}{lc}
    \hline
         Itemset & Support \\
    \hline     
         Jacket & 2 \\
         Outwear & 3 \\
         Clothes & 4 \\
         Shoes & 2 \\
         Hiking Boots & 2 \\
         Footwear & 4 \\
         Outwear, Hiking Boots & 2\\
         Clothes, Hiking Boots & 2\\
         Outwear, Footwear & 2\\
         Clothes, Footwear & 2\\
    \hline 
    \end{tabular}
  \end{subfigure}
  \begin{subfigure}[b]{0.5\linewidth}
    \centering
    \caption{Association rules} 
    \label{fig7:d}
      \begin{tabular}{lcc}
    \hline
         Rule & Support & Confidence \\
        \hline
         Outwear $\to$ Hiking Boots & 1/3 & 2/3\\
         Outwear $\to$ Footwear & 1/3 & 2/3\\
         Hiking Boots $\to$ Outwear & 1/3 & 1 \\
         Hiking Boots $\to$ Clothes & 1/3 & 1 \\
    \hline
    \end{tabular}
    
  \end{subfigure} 
  \caption{A database of transactions, taxonomies and extracted rules~\cite{Srikant:95,Srikant:97}}
  \label{fig:genattr_setting} 
\end{figure}

To the best of our knowledge the problem of mining generalised association rules was first introduced around 1995 in~\cite{Srikant:95,Srikant:97}, and rephrased as follows: Given a large database of transactions, where each transaction consists of a set of items, and a taxonomy (is-a hierarchy) on the items, the goal is to find associations between items at any level of the taxonomy. For example, with a taxonomy that says that \texttt{jackets} is-a \texttt{outerwear} and \texttt{outerwear} is-a \texttt{clothes}, we may infer a rule that ``people who buy \texttt{outerwear} tend to buy \texttt{shoes}''. This rule may hold even if rules that ``people who buy \texttt{jackets} tend to buy \texttt{shoes}'', and ``people who buy \texttt{clothes} tend to buy \texttt{shoes}'' do not hold. (See Fig.~\ref{fig:genattr_setting})


A generalised association rule is a (partial) implication $X \to Y$, where $X, Y$ are disjoint itemsets and no item in $Y$ is a generalisation of any item in $X$~\cite{Srikant:95,Srikant:97}. We adopt the following notation: 
$\mathcal{I}=\{i_1,i_2, \cdots, i_m \}$ is a set of items and $\mathcal{D} = \{t_1, t_2, \cdots, t_n\}$ a set of transactions. Each transaction $t \in \mathcal{D}$ is a subset of items $\mathcal{I}$. 
Let $\mathcal{T}$ be a set of taxonomies (i.e directed acyclic graph on items and generalised items). 
We denote by $(\mathcal{T},\le)$ its transitive closure. The elements of $\mathcal{T}$ are called ``general items''. A transaction $t$ \emph{supports} an item $x$ (resp. a general item $y$) if $x$ is in $t$ (resp. $y$ is a generalisation of an item $x$ in $t$). 
A set of transactions $T$ \emph{supports} an itemset $X\subseteq \mathcal{I}$ if $T$ supports every item in $X$.



In FCA setting, we build a generalised context $(\mathcal{D},\mathcal{I}\cup \mathcal{T}, I)$, where the set  of objects, $\mathcal{D}$, 
is the set of transactions (strictly speaking transaction-ID), and the set of attributes, $M=\mathcal{I}\cup \mathcal{T}$, contains all items ($\mathcal{I}$) and general items ($\mathcal{T}$).  The incidence relation $I \subseteq \mathcal{D} \times M$ is defined by
		\[
		tIm \iff 
		\begin{cases}
		m\in\mathcal{I} \text{ and } m\in t \\
		m\in\mathcal{T} \text{ and } \exists n\in\mathcal{I}, n\in t \text{ and } n\le m.
		\end{cases}    
		\]

Below is the context associated to the example on Figure~\ref{fig:genattr_setting}. 
\begin{center}
\setlength{\tabcolsep}{3pt}
\begin{tabular}{|c|c|c|c|c|c||c|c|c|}
\hline
 & Shirt & Jacket & Hiking Boots & Ski Pants & Shoes & Outerwear & Clothes & Footwear \\ \hline
100 & $\times$ &&&&&&$\times$& \\ \hline
200 & & $\times$ & $\times$ &&&$\times$&$\times$&$\times$ \\ \hline 
300 & & & $\times$ & $\times$ &&$\times$&$\times$&$\times$ \\ \hline 
400 &&&& & $\times$ &&&$\times$\\ \hline
500 &&&& & $\times$ &&&$\times$ \\ \hline
600 && $\times$ &&&& $\times$ & $\times$ & \\ \hline
\end{tabular}
\end{center}

The basic interestingness measures for a generalised rule $X \to Y$ are support and confidence (see association rules in~Fig.~\ref{fig:genattr_setting} (d)). Its \emph{support} $supp(X \to Y)$ is defined as $\frac{|(X \cup Y)'|}{|\mathcal D|}$, while its \emph{confidence} $conf(X \to Y)$ is $\frac{|(X \cup Y)'|}{|X'|}$.   

For some applications, it would make sense to work only with the subcontext $(\mathcal{D},\mathcal{T}, I \cap \mathcal{D} \times \mathcal{T})$ instead of $(\mathcal{D},\mathcal{I}\cup \mathcal{T}, I)$, for example if the goal is to reduce the number of attributes, concepts or rules. Sometimes, there is no taxon available to suggest that considered attributes should be put together. However, we can extend the used taxonomy, i.e. put some attributes together in a proper taxon, and decide when an object satisfies the grouped attributes.


\subsection{Generalising scenarios}
Let $\mathbb{K}:=(G,M,I)$ be a context. The attributes of $\mathbb{K}$ can be grouped to form another set of attributes, namely $S$, whose elements are called \textbf{generalised} attributes. For example, in basket market analysis, items (products) can be generalised into product lines and then product categories, and even customers may be generalised to groups according to specific features (e.g., income, education). This replaces $(G,M,I)$ with a context $(G,S,J)$ where $S$ can be seen as an index set such that $\{m_s\mid s\in S\}$ covers $M$. 
 How to define the incidence relation $J$, is domain dependent. Let us consider several cases below~\cite{Kwuida:2009,Kwuida:2012,Kwuida:2014}:

\begin{itemize}
	\item[$(\exists)$] $gJs:\iff \exists m\in s,\, g\I m$. \ When companies are described by the locations of their branches then cities can be grouped to regions or states. 
	A company $g$ operates in a state $s$ if $g$ has a branch in a city $m$ which is in $s$. 
	\item[$(\forall)$] $gJs:\iff \forall m\in s,\, g\I m$. \ For exams with several components (e.g. written, oral, and thesis), we might require students to pass all components in order to succeed.
	\item[$(\alpha\%)$] $gJs:\iff \frac{|\{m\in s\ \mid\ g I m\}|}{|s|}\geq\alpha_s$ with $\alpha_s$ a threshold. \ In the case of exams discussed above, we could require students to pass just some parts, defined by a threshold.
\end{itemize}

Similarly, objects can also be put together to get ``generalised objects''. In \cite{Prediger:2003} the author described \emph{general on objects} as 
 classes of individual objects that are considered to be extents of concepts
of a formal context. In that paper, different contexts with general objects are defined and their conceptual structure and relation to other contexts is analysed with FCA methods. Generalisation on both objects and attributes can be carried out with the combinations below, with $A \subseteq G$ and $B \subseteq M$:

\begin{enumerate}
	\item $AJB$ iff $\exists a\in A$, $\exists b\in B$ such that $a\I b$ (i.e. some objects from $A$ are in relation with some attributes in $B$);
	\item $AJB$ iff $\forall a\in A$, $\forall b\in B\  a\I b$ (i.e. each object in $A$ has all attributes in $B$);
	\item $AJB$ iff $\forall a\in A$, $\exists b\in B$ such that $a\I b$ (i.e. each object in $A$ has at least one attribute in $B$);
	\item $AJB$ iff $\exists b\in B$ such that $\forall a\in A$  $a\I b$ (i.e. an attribute in $B$ is satisfied by all objects of $A$);
	\item $AJB$ iff $\forall b\in B$, $\exists a\in A$ such that $a\I b$ ( i.e. each property in $B$ is satisfied by an object of $A$);
	\item $AJB$ iff $\exists a\in A$ such that  $\forall b\in B\ a\I b$ (i.e. an object in $A$ has all attributes in $B$);
	\item $AJB$ iff $\frac{\left|\{a\in A\mid \frac{|\{b\in B\mid a\I b\}|}{|B|}\geq \beta_{B}\} \right|}{|A|}\geq\alpha_A$ (i.e. at least $\alpha_{A}\%$ of objects in $A$ have each at least $\beta_{B}\%$ of the attributes in $B$);
	\item $AJB$ iff $\frac{\left|\left\{b\in B\mid
		\frac{|\{a\in A\mid a\I b\}|}{|A|}\geq
		\alpha_{A}\right\} \right|}{|B|}\geq\beta_B$ (i.e. at least $\beta_{B}\%$ of attributes in $B$ belong altogether to at least $\alpha_{A}\%$ of objects in the group $A$);
	\item $AJB$ iff $\frac{|A\times B\cap I|}{|A\times B|}\geq \alpha$ (i.e. the density of the rectangle $A\times B$ is at least $\alpha$).
\end{enumerate}

%
\subsection{Generalisation and extracted patterns}
After analysing several generalisation cases, including simultaneous generalisations on both objects and attributes as above, the next step is to look at the extracted patterns. From contexts, knowledge is usually extracted in terms of clusters and rules. When dealing with generalised attributes or objects, we coin the term ``generalised'' to all patterns extracted. An immediate task is to compare knowledge gained after generalising with those from the initial context.

New and interesting rules as seen in Figure~\ref{fig:genattr_setting} can be discovered~\cite{Srikant:95,Srikant:97}. 
Experiments have shown that the number of extracted patterns quite often decreases. Formal investigations are been carried out to compare these numbers. For $\forall$-generalisations,  the number of concepts does not increase~\cite{Kwuida:2014}. But for $\exists$-generalisations, the size can actually increase~\cite{Kwuida:2009,Kwuida:2012,Kwuida:2014,Kwuida:2017,Kwuida:2020}.

In~\cite{Belohlavek:2014} the authors propose a method to control the structure of concept lattices derived from Boolean data by specifying granularity levels of attributes. Here a taxonomy is already available, given by the granularity of the attributes. They suggest that granularity levels should be chosen by a user based on his expertise and experimentation with the data. If the resulting formal concepts are too specific and there is a large number of them, the user can choose to use a coarser level of granularity. The resulting formal concepts are then less specific and can be seen as resulting from a zoom-out. Similarly, one may perform a zoom-in to obtain finer, more specific formal concepts. Through all these precautions, the number of concepts can still increase when attributes are coarser: ``The issue of when attribute coarsening results in an increase in the number of formal concepts needs a further examination, as well as the possibility of informing automatically a user who is selecting a new level of granularity that the new level results in an increase in the number of concepts.''~\cite{Belohlavek:2014}

In~\cite{Kwuida:2020} a more succinct analysis of $\exists$-generalisations presents a family of contexts where generalising two attributes results in an exponential increase in the number of concepts. An example of such context is given in the Table~\ref{tab:S6_7} (left).

\begin{table}[ht]
   \caption{A formal context (left) and its $\exists$-generalisation that puts $m_1$ and $m_2$ together. The number of concepts increases from 48 to 64, i.e. by 16.}
    \label{tab:S6_7}
    \centering
\begin{tabular}{|l|l|l|l|l|l|l|l|l|} \hline
& $1$ & $2$ & $3$ & $4$ & $5$ & $6$ & $m_{1}$ & $m_{2}$ \\ \hline
$1$ & & $\times $ & $\times $ & $\times $ &$\times$ & $\times$ & $\times$ & \\ \hline
$2$ & $\times $ &  & $\times $ & $\times $ & $\times $ &$\times$ & $\times$ & $\times$ \\ \hline
$3$ & $\times $ & $\times$ & & $\times $ & $\times $ &$\times$ & $\times$ & $\times$ \\ \hline
$4$ & $\times $ & $\times$ & $\times$ & & $\times $ &$\times$ & $\times$ & $\times$ \\ \hline
$5$ & $\times $ & $\times$ & $\times$ & $\times $ & & $\times$ & $\times$ & $\times$ \\ \hline
$6$ & $\times $ & $\times$ & $\times$ & $\times $ &$\times$ & & & $\times$ \\ \hline
$g_{1}$ & $\times $ & $\times $ & $\times$ & $\times $ & $\times$ & $\times$ & & \\ \hline
\end{tabular}
\quad $\Longrightarrow$\quad 
\begin{tabular}{|l|l|l|l|l|l|l|l|} \hline
& $1$ & $2$ & $3$ & $4$ & $5$ & $6$ & $m_{12}$ \\ \hline
$1$ & & $\times $ & $\times $ & $\times $ &$\times$ & $\times$ & $\times$ \\ \hline
$2$ & $\times $ &  & $\times $ & $\times $ & $\times $ &$\times$ & $\times$  \\ \hline
$3$ & $\times $ & $\times$ & & $\times $ & $\times $ &$\times$ & $\times$ \\ \hline
$4$ & $\times $ & $\times$ & $\times$ & & $\times $ &$\times$ & $\times$  \\ \hline
$5$ & $\times $ & $\times$ & $\times$ & $\times $ & & $\times$ & $\times$  \\ \hline
$6$ & $\times $ & $\times$ & $\times$ & $\times $ &$\times$ & & $\times$ \\ \hline
$g_{1}$ & $\times $ & $\times $ & $\times$ & $\times $ & $\times$ & $\times$ & \\ \hline
\end{tabular}
\end{table}
Putting together some attributes does not always reduce the number of extracted patterns. It's therefore interesting to get measures that suggest which attributes can be put together, in the absence of a taxonomy. The goal would be to not increase the size of extracted patterns. 

\subsection{Similarity and existential generalisations}
This section presents investigations on the use of certain similarity measures in generalising attributes. 
	A \emph{similarity measure} on a set $M$ of attributes is a function $S:M\times M\to \RR$  such that for all $m_1,m_2$ in $M$,
	\begin{itemize}
		\item[(i)]  $S(m_1,m_2)\geq 0$, \hfill \textit{positivity}
		\item[(ii)] $S(m_1,m_2)=S(m_2,m_1)$ \hfill  \textit{symmetry}
		\item[(iii)] $S(m_1,m_1) \geq S(m_1,m_2)$ \hfill \textit{maximality}
	\end{itemize}
We say that $S$ is \emph{compatible} with generalising attributes if whenever $m_1, m_2$ are more similar than $m_3,m_4$, then putting  $m_1,m_2$ together should not lead to more concepts than putting  $m_3,m_4$ together does. 
To give the formula for some known similarity measures that could be of interest in FCA setting, we adopt the following notation for $m_1,m_2$ attributes in $\KK$:   
\[a=|m_1'\cap m_2'|,\quad d=|m_1'\Delta m_2'|, \quad b=|m_1'\setminus m_2'|,\quad c=|m_2'\setminus m_1'|.\]

\begin{table}[htbp]
     \caption{Some similarity measures relevant in FCA}
     \label{tab:similarities}
     \centering
\begin{small}
\renewcommand{\arraystretch}{2.5} \tabcolsep=4pt
\begin{tabular}{|lr||lr|} \hline
Name & Formula & 	Name & Formula \\ \hline
Jaccard (Jc)& $\dfrac{a}{a+b+c}$ & Sneath/Sokal (SS$_1$) & $\dfrac{2(a+d)}{2(a+d)+b+c}$  \\ 
Dice (Di) & $\dfrac{2a}{2a+b+c}$ & Sneath/Sokal (SS$_2$) & $\dfrac{0.5a}{0.5a+b+c}$ \\ 
Sorensen (So) & $\dfrac{4a}{4a+b+c}$ & Sokal/Michener (SM) & $\dfrac{a+d}{a+d+b+c}$\\ 
Anderberg (An) & $\dfrac{8a}{8a+b+c}$ & Rogers/Tanimoto (RT) & $\dfrac{0.5(a+d)}{0.5(a+d)+b+c}$ \\ 
Orchiai (Or) &  $\dfrac{a}{\sqrt{(a+b)(a+c}}$ & Russels/Rao (RR) & $\dfrac{a}{a+d+b+c}$ \\ 
Kulczynski (Ku) &  $\dfrac{0.5a}{a+b}+\dfrac{0.5a}{a+c}$ & Yule/Kendall (YK) & $\dfrac{ad}{ad+bc}$ \\ \hline
\end{tabular}
\end{small}

 \end{table}
%

\noindent
For the context left in Table~\ref{tab:S6_7}, we have computed $S(m_1,x),\  x=1,\ldots , 6, m_2$. Although $m_1$ is more similar to $m_2$ than any attribute $i<6$, putting $m_1$ and $m_2$ together increases the number of concepts. Note that putting $m_1$ and $6$ together is equivalent to removing $m_1$ from the context, and thus, reduces the number of concepts. 

\begin{table}[ht]
\caption{The values of considered similarity measures $S(m,i)$}\label{tab:sim_val}
\def\arraystretch{1.2}\tabcolsep=4pt
\centering
\begin{tabular}{|c  r  r r r r r r r r r r|}
\hline
& Jc & Di & So & An & SS$_2$ & Ku & Or & SM & RT & SS$_1$ & RR  \\
\hline
$i\in S_5$ & 0.57 & 0.80 & 0.89 & 0.94 & 0.50 & 0.80 & 0.80 & 0.71 & 0.56 & 0.83 & 0.57 \\
$i=6$ & 0.83 & 0.91 & 0.95 & 0.97 & 0.71 & 0.92 & 0.91 & 0.75 & 0.75 & 0.92 & 0.71 \\
$i=m_2$ & 0.67 & 0.80 & 0.89 & 0.94 & 0.50 & 0.80 & 0.80 & 0.71 & 0.56 & 0.83 & 0.57 \\ 	\hline
\end{tabular}
\end{table}

Let $\KK$ be a context $(G,M,I)$ with $a,b\in M$ and 
$\KK_{00}$ be its subcontext without $a,b$. Below, $\Ext(\KK_{00}$) means all the extents of concepts of the context $\KK_{00}$. In order to describe the increase in the number of concepts after putting $a,b$ together, we set
\begin{align*}
\mathcal{H}(a)&:=
\left\{A\cap a^\prime \mid A\in\Ext(\KK_{00}) \text{ and }  A\cap a^\prime \notin\Ext(\KK_{00})\right\} \\
\mathcal{H}(b)&:=
\left\{A\cap b^\prime \mid A\in\Ext(\KK_{00}) \text{ and }  A\cap b^\prime \notin\Ext(\KK_{00})\right\} \\
\mathcal{H}(a\cup b)&:=
\left\{A\cap (a^\prime\cup b^\prime)\mid A\in\Ext(\KK_{00}),\  A\cap(a^\prime\cup b^\prime)\notin\Ext(\KK_{00})\right\} \\
\mathcal{H}(a\cap b)&:=
\left\{A\cap (a^\prime\cap b^\prime) \mid A\in\Ext(\KK_{00}),\ A\cap(a^\prime\cap b^\prime)\notin\Ext(\KK_{00})\right\}.
\end{align*}

\noindent
The following proposition shows that the increase can be exponential. 
\begin{theorem}[\cite{Kwuida:2020}]
	Let $(G,M,\I)$ be an attribute reduced context and $a, b$ be  two attributes such that their generalisation $s=a\cup b$ increases the size of the concept lattice.  Then
$|\frak{B}(G,M,I)| = |\frak{B}(G,M\setminus\{a,b\},I \subseteq G \times M\setminus\{a,b\} )|+|\cal{H}(a,b)|$, with
		\begin{align*}
		|\cal{H}(a,b)|&=|\cal{H}(a)\cup \cal{H}(b)\cup \cal{H}(a\cap b)| \le 2^{|a'|+|b'|}-2^{|a'|}-2^{|b'|}+1.
		\end{align*}
This upper bound can be reached. 
\end{theorem}
The difference $|\cal{H}(a,b)|$ is then used to define a compatible similarity measure. We set 

$\psi(a,b) := |\cal{H}(a\cup b)|-|\cal{H}(a,b)|$,\quad 
 $\delta(a,b):=
\begin{cases}
1 & \text{ if \ }\psi (a,b)\leq 0 \\
0 & \text{ else}%
\end{cases}%
$,  and define

$S(a,b):=\dfrac{1+\delta(a,b)}{2}-\dfrac{|\psi(a,b)|}{2n_0}$ with $n_0=\max\{\psi(x,y)\mid x,y\in M\}$. Then

\begin{theorem}[\cite{Kuitche:2018}]
$S$ is a similarity measure compatible with the generalisation.
\[S(a,b)\ge \frac{1}{2} \iff \psi(a,b)\le 0\]
\end{theorem}

\section{Conclusion}\label{sec:Concl}

The first two parts contain a concise summary of the usage of Shapley values from Cooperative Game Theory for interpretable concept-based learning in the FCA playground with its connection to Data Mining formulations. We omitted results related to algorithms and their computational complexity since they deserve a separate detailed treatment. 

The lessons drawn from the ranking attributes in JSM classification hypotheses and those in the intents of stable concepts show that valuation functions should be customised and are not necessarily zero-one-valued.
This is an argument towards that of Shapley values approach requires specification depending on the model (or type of patterns) and thus only conditionally is model-agnostic. The other lesson is about the usage of Shapley values for pattern attribution concerning their contribution interestingness measures like stability or robustness. 

The third part is devoted to attribute aggregation by similarity, which may help to apply interpretable techniques to larger sets of attributes or bring additional aspects to interpretability with the help of domain taxonomies. 
The desirable property of similarity measures to provide compatible generalisation helps to reduce the number of output concepts or JSM-hypotheses as well. 
The connection between attribute similarity measures and Shapley interaction values~\cite{Lundberg:2017}, when the interaction of two or more attributes on the model prediction is studied, is also of interest. 

In addition to algorithmic issues, we would like to mention two more directions of future studies. The first one lies in the interpretability by means of Boolean matrix factorisation (decomposition), which was used for dimensionality reduction with explainable Boolean factors (formal concepts)~\cite{Belohlavek:2010} or interpretable ``taste communities'' identification in collaborative filtering~\cite{Ignatov:2014b}. In this case, we are transitioned from the importance of attributes to attribution of factors. The second one is a closely related aspect to interpretability called fairness~\cite{Alves:2020}, where, for example, certain attributes of individuals should not influence much to the model prediction (disability, ethnicity, gender, etc.). 

\subsubsection*{Acknowledgements.} The study was implemented in the framework of the Basic Research Program at the National Research University Higher School of Economics and funded by the Russian Academic Excellence Project '5-100'. The second author was also supported by Russian Science Foundation under grant 17-11-01276 at St. Petersburg Department of Steklov Mathematical Institute of Russian Academy of Sciences, Russia. The second author would like to thank Fuad Aleskerov, Alexei Zakharov, and Shlomo Weber  for the inspirational lectures on Collective Choice and Voting Theory.

\bibliographystyle{splncs04}
\bibliography{imlbib}

\begin{thebibliography}{10}
\providecommand{\url}[1]{\texttt{#1}}
\providecommand{\urlprefix}{URL }
\providecommand{\doi}[1]{https://doi.org/#1}

\bibitem{Agrawal:93}
Agrawal, R., Imielinski, T., Swami, A.N.: Mining association rules between sets
  of items in large databases. In: Buneman, P., Jajodia, S. (eds.) Proceedings
  of the 1993 {ACM} {SIGMOD} International Conference on Management of Data,
  Washington, DC, USA, May 26-28, 1993. pp. 207--216. {ACM} Press (1993)

\bibitem{Alves:2020}
Alves, G., Bhargava, V., Couceiro, M., Napoli, A.: Making {ML} models fairer
  through explanations: the case of limeout. CoRR  \textbf{abs/2011.00603}
  (2020)

\bibitem{Belohlavek:2014}
Belohl{\'{a}}vek, R., Baets, B.D., Konecny, J.: Granularity of attributes in
  formal concept analysis. Inf. Sci.  \textbf{260},  149--170 (2014)

\bibitem{Belohlavek:2009}
Belohl{\'{a}}vek, R., Baets, B.D., Outrata, J., Vychodil, V.: Inducing decision
  trees via concept lattices. Int. J. Gen. Syst.  \textbf{38}(4),  455--467
  (2009)

\bibitem{Belohlavek:2010}
Belohl{\'{a}}vek, R., Vychodil, V.: Discovery of optimal factors in binary data
  via a novel method of matrix decomposition. J. Comput. Syst. Sci.
  \textbf{76}(1),  3--20 (2010)

\bibitem{Bocharov:2016}
Bocharov, A., Gnatyshak, D., Ignatov, D.I., Mirkin, B.G., Shestakov, A.: A
  lattice-based consensus clustering algorithm. In: Huchard, M., Kuznetsov,
  S.O. (eds.) Proceedings of the Thirteenth International Conference on Concept
  Lattices and Their Applications, Moscow, Russia, July 18-22, 2016. {CEUR}
  Workshop Proceedings, vol.~1624, pp. 45--56. CEUR-WS.org (2016)

\bibitem{Carpineto:96}
Carpineto, C., Romano, G.: A lattice conceptual clustering system and its
  application to browsing retrieval. Mach. Learn.  \textbf{24}(2),  95--122
  (1996)

\bibitem{Caruana:2020}
Caruana, R., Lundberg, S., Ribeiro, M.T., Nori, H., Jenkins, S.: Intelligible
  and explainable machine learning: Best practices and practical challenges.
  In: Gupta, R., Liu, Y., Tang, J., Prakash, B.A. (eds.) {KDD} '20: The 26th
  {ACM} {SIGKDD} Conference on Knowledge Discovery and Data Mining, Virtual
  Event, CA, USA, August 23-27, 2020. pp. 3511--3512. {ACM} (2020)

\bibitem{Eklund:2012}
Eklund, P.W., Ducrou, J., Dau, F.: Concept similarity and related categories in
  information retrieval using {Formal Concept Analysis}. Int. J. Gen. Syst.
  \textbf{41}(8),  826--846 (2012)

\bibitem{Fayyad:96}
Fayyad, U.M., Piatetsky{-}Shapiro, G., Smyth, P.: From data mining to knowledge
  discovery in databases. {AI} Magazine  \textbf{17}(3),  37--54 (1996)

\bibitem{Finn:1983}
Finn, V.: {On Machine-oriented Formalization of Plausible Reasoning in
  F.Bacon-J.S.Mill Style}. Semiotika i Informatika (20),  35--101 (1983), (in
  Russian)

\bibitem{Ganter:2003}
Ganter, B., Kuznetsov, S.O.: {Hypotheses and Version Spaces}. In: de~Moor, A.,
  Lex, W., Ganter, B. (eds.) Conceptual Structures for Knowledge Creation and
  Communication, 11th International Conference on Conceptual Structures, {ICCS}
  2003, Proceedings. LNCS, vol.~2746, pp. 83--95. Springer (2003)

\bibitem{Ganter:2008}
Ganter, B., Kuznetsov, S.O.: {Scale Coarsening as Feature Selection}. In:
  Medina, R., Obiedkov, S. (eds.) Formal Concept Analysis. pp. 217--228.
  Springer Berlin Heidelberg (2008)

\bibitem{Ganter:2016}
Ganter, B., Obiedkov, S.A.: Conceptual Exploration. Springer (2016)

\bibitem{Ganter:1999}
Ganter, B., Wille, R.: Formal Concept Analysis - Mathematical Foundations.
  Springer (1999)

\bibitem{Goodfellow:2016}
Goodfellow, I.J., Bengio, Y., Courville, A.C.: Deep Learning. Adaptive
  computation and machine learning, {MIT} Press (2016)

\bibitem{Hajek:1966}
H{\'a}jek, P., Havel, I., Chytil, M.: {The GUHA method of automatic hypotheses
  determination}. Computing  \textbf{1}(4),  293--308 (1966)

\bibitem{Ignatov:2014a}
Ignatov, D.I.: Introduction to formal concept analysis and its applications in
  information retrieval and related fields. In: Braslavski, P., Karpov, N.,
  Worring, M., Volkovich, Y., Ignatov, D.I. (eds.) Information Retrieval - 8th
  Russian Summer School, RuSSIR 2014, Nizhniy, Novgorod, Russia, August 18-22,
  2014, Revised Selected Papers. Communications in Computer and Information
  Science, vol.~505, pp. 42--141. Springer (2014)

\bibitem{Ignatov:2012}
Ignatov, D.I., Kuznetsov, S.O., Poelmans, J.: {Concept-Based Biclustering for
  Internet Advertisement}. In: 12th {IEEE} International Conference on Data
  Mining Workshops, {ICDM} Workshops, Brussels, Belgium, December 10, 2012. pp.
  123--130 (2012)

\bibitem{Ignatov:2020a}
Ignatov, D.I., Kwuida, L.: Interpretable concept-based classification with
  shapley values. In: Alam, M., Braun, T., Yun, B. (eds.) Ontologies and
  Concepts in Mind and Machine - 25th International Conference on Conceptual
  Structures, {ICCS} 2020, Bolzano, Italy, September 18-20, 2020, Proceedings.
  Lecture Notes in Computer Science, vol. 12277, pp. 90--102. Springer (2020)

\bibitem{Ignatov:2020b}
Ignatov, D.I., Kwuida, L.: Shapley and banzhaf vectors of a formal concept. In:
  Valverde{-}Albacete, F.J., Trnecka, M. (eds.) Proceedings of the Fifthteenth
  International Conference on Concept Lattices and Their Applications, Tallinn,
  Estonia, June 29-July 1, 2020. {CEUR} Workshop Proceedings, vol.~2668, pp.
  259--271. CEUR-WS.org (2020)

\bibitem{Ignatov:2014b}
Ignatov, D.I., Nenova, E., Konstantinova, N., Konstantinov, A.V.: {Boolean
  Matrix Factorisation for Collaborative Filtering: An FCA-Based Approach}. In:
  Agre, G., Hitzler, P., Krisnadhi, A.A., Kuznetsov, S.O. (eds.) Artificial
  Intelligence: Methodology, Systems, and Applications - 16th International
  Conference, {AIMSA} 2014, Varna, Bulgaria, September 11-13, 2014.
  Proceedings. Lecture Notes in Computer Science, vol.~8722, pp. 47--58.
  Springer (2014)

\bibitem{Mill:1843}
John, S.: Mill, A System of Logic, Ratiocinative and Inductive, Being a
  Connected View of the Principles of Evidence and the Methods of Scientific
  Investigation. Longmans, Green, and Co., London (1843)

\bibitem{Kadyrov:2019}
Kadyrov, T., Ignatov, D.I.: Attribution of customers’ actions based on
  machine learning approach. In: Proceedings of the Fifth Workshop on
  Experimental Economics and Machine Learning co-located with the Seventh
  International Conference on Applied Research in Economics (iCare7), Perm,
  Russia, September 26, 2019. CEUR-ws, vol. Vol-2479, pp. 77--88 (2019)

\bibitem{Kashnitsky:2016}
Kashnitsky, Y., Kuznetsov, S.O.: {Global Optimization in Learning with
  Important Data: an FCA-Based Approach}. In: Huchard, M., Kuznetsov, S.O.
  (eds.) Proceedings of the Thirteenth International Conference on Concept
  Lattices and Their Applications, Moscow, Russia, July 18-22, 2016. {CEUR}
  Workshop Proceedings, vol.~1624, pp. 189--201. CEUR-WS.org (2016)

\bibitem{Kaur:2020}
Kaur, H., Nori, H., Jenkins, S., Caruana, R., Wallach, H.M., Vaughan, J.W.:
  Interpreting interpretability: Understanding data scientists' use of
  interpretability tools for machine learning. In: Bernhaupt, R., Mueller,
  F.F., Verweij, D., Andres, J., McGrenere, J., Cockburn, A., Avellino, I.,
  Goguey, A., Bj{\o}n, P., Zhao, S., Samson, B.P., Kocielnik, R. (eds.) {CHI}
  '20: {CHI} Conference on Human Factors in Computing Systems, Honolulu, HI,
  USA, April 25-30, 2020. pp. 1--14. {ACM} (2020)

\bibitem{Kaytoue:2014}
Kaytoue, M., Kuznetsov, S.O., Macko, J., Napoli, A.: Biclustering meets triadic
  concept analysis. Ann. Math. Artif. Intell.  \textbf{70}(1-2),  55--79 (2014)

\bibitem{Konecny:2017}
Konecny, J.: On attribute reduction in concept lattices: Methods based on
  discernibility matrix are outperformed by basic clarification and reduction.
  Inf. Sci.  \textbf{415},  199--212 (2017)

\bibitem{Konecny:2018}
Konecny, J., Krajca, P.: On attribute reduction in concept lattices:
  Experimental evaluation shows discernibility matrix based methods
  inefficient. Inf. Sci.  \textbf{467},  431--445 (2018)

\bibitem{Kuitche:2018}
Kuitch{\'{e}}, R.S., Temgoua, R.E.A., Kwuida, L.: A similarity measure to
  generalize attributes. In: Ignatov, D.I., Nourine, L. (eds.) Proceedings of
  the Fourteenth International Conference on Concept Lattices and Their
  Applications, {CLA} 2018, Olomouc, Czech Republic, June 12-14, 2018. {CEUR}
  Workshop Proceedings, vol.~2123, pp. 141--152. CEUR-WS.org (2018)

\bibitem{Kuznetsov:2004}
Kuznetsov, S.O.: {Machine Learning and Formal Concept Analysis}. In: {ICFCA}
  2004. pp. 287--312 (2004)

\bibitem{Kuznetsov:2005}
Kuznetsov, S.O.: {Galois Connections in Data Analysis: Contributions from the
  Soviet Era and Modern Russian Research}. In: Ganter, B., Stumme, G., Wille,
  R. (eds.) Formal Concept Analysis, Foundations and Applications. LNCS,
  vol.~3626, pp. 196--225. Springer (2005)

\bibitem{Kuznetsov:2007a}
Kuznetsov, S.O.: On stability of a formal concept. Ann. Math. Artif. Intell.
  \textbf{49}(1-4),  101--115 (2007)

\bibitem{Kuznetsov:2018}
Kuznetsov, S.O., Makhalova, T.P.: On interestingness measures of formal
  concepts. Inf. Sci.  \textbf{442-443},  202--219 (2018)

\bibitem{Kuznetsov:2017}
Kuznetsov, S.O., Makhazhanov, N., Ushakov, M.: On neural network architecture
  based on concept lattices. In: Kryszkiewicz, M., Appice, A., Slezak, D.,
  Rybinski, H., Skowron, A., Ras, Z.W. (eds.) Foundations of Intelligent
  Systems. pp. 653--663. Springer International Publishing, Cham (2017)

\bibitem{Kuznetsov:2013}
Kuznetsov, S.O., Poelmans, J.: Knowledge representation and processing with
  formal concept analysis. Wiley Interdiscip. Rev. Data Min. Knowl. Discov.
  \textbf{3}(3),  200--215 (2013)

\bibitem{Kuznetsov:1991}
Kuznetsov, S.: Jsm-method as a machine learning method. Method. Itogi Nauki i
  Tekhniki, ser. Informatika (15),  17--53 (1991), (in Russian)

\bibitem{Kuznetsov:1990}
Kuznetsov, S.: Stability as an estimate of the degree of substantiation of
  hypotheses derived on the basis of operational similarity. Nauchn. Tekh. Inf.
  Ser. 2 (12),  217--29 (1991), (in Russian)

\bibitem{Kuznetsov:1996}
Kuznetsov, S.: Mathematical aspects of concept analysis. Journal of
  Mathematical Science  \textbf{80}(2),  1654--1698 (1996)

\bibitem{Kwuida:2017}
Kwuida, L., Kuitch{\'e}, R., Temgoua, R.: On the size of $\exists$-generalized
  concepts. ArXiv:1709.08060  (2017)

\bibitem{Kwuida:2020}
Kwuida, L., Kuitch{\'{e}}, R.S., Temgoua, R.E.A.: On the size of
  {\(\exists\)}-generalized concept lattices. Discret. Appl. Math.
  \textbf{273},  205--216 (2020)

\bibitem{Kwuida:2014}
Kwuida, L., Missaoui, R., Balamane, A., Vaillancourt, J.: Generalized pattern
  extraction from concept lattices. Ann. Math. Artif. Intell.
  \textbf{72}(1-2),  151--168 (2014)

\bibitem{Kwuida:2009}
Kwuida, L., Missaoui, R., Boumedjout, L., Vaillancourt, J.: Mining generalized
  patterns from large databases using ontologies. ArXiv:0905.4713  (2009)

\bibitem{Kwuida:2012}
Kwuida, L., Missaoui, R., Vaillancourt, J.: Using taxonomies on objects and
  attributes to discover generalized patterns. In: Szathmary, L., Priss, U.
  (eds.) Proceedings of The Ninth International Conference on Concept Lattices
  and Their Applications, Fuengirola (M{\'{a}}laga), Spain, October 11-14,
  2012. {CEUR} Workshop Proceedings, vol.~972, pp. 327--338. CEUR-WS.org (2012)

\bibitem{Lakhal:2005}
Lakhal, L., Stumme, G.: Efficient mining of association rules based on formal
  concept analysis. In: Ganter, B., Stumme, G., Wille, R. (eds.) Formal Concept
  Analysis, Foundations and Applications. Lecture Notes in Computer Science,
  vol.~3626, pp. 180--195. Springer (2005)

\bibitem{Lundberg:2017}
Lundberg, S.M., Lee, S.I.: {A Unified Approach to Interpreting Model
  Predictions}. In: et~al., I.G. (ed.) Advances in Neural Information
  Processing Systems 30, pp. 4765--4774. Curran Associates, Inc. (2017)

\bibitem{Luxenburger:1991}
Luxenburger, M.: Implications partielles dans un contexte. Math{\'e}matiques et
  Sciences Humaines  \textbf{113},  35--55 (1991)

\bibitem{Mirkin:1996}
Mirkin, B.: Mathematical Classification and Clustering. Kluwer Academic
  Publishers (1996)

\bibitem{Mitchell:1977}
Mitchell, T.M.: {Version Spaces: {A} Candidate Elimination Approach to Rule
  Learning}. In: Reddy, R. (ed.) Proceedings of the 5th International Joint
  Conference on Artificial Intelligence. 1977. pp. 305--310. William Kaufmann
  (1977)

\bibitem{Molnar:2019}
Molnar, C.: {Interpretable Machine Learning} (2019),
  \url{https://christophm.github.io/interpretable-ml-book/}

\bibitem{Pasquier:1999}
Pasquier, N., Bastide, Y., Taouil, R., Lakhal, L.: Efficient mining of
  association rules using closed itemset lattices. Information Systems
  \textbf{24}(1),  25--46 (1999)

\bibitem{Poelmans:2013a}
Poelmans, J., Ignatov, D.I., Kuznetsov, S.O., Dedene, G.: Formal concept
  analysis in knowledge processing: {A} survey on applications. Expert Syst.
  Appl.  \textbf{40}(16),  6538--6560 (2013)

\bibitem{Poelmans:2013b}
Poelmans, J., Kuznetsov, S.O., Ignatov, D.I., Dedene, G.: Formal concept
  analysis in knowledge processing: {A} survey on models and techniques. Expert
  Syst. Appl.  \textbf{40}(16),  6601--6623 (2013)

\bibitem{Prediger:2003}
Prediger, S.: Formal concept analysis for general objects. Discret. Appl. Math.
   \textbf{127}(2),  337--355 (2003)

\bibitem{Priss:2011}
Priss, U., Old, L.J.: Data weeding techniques applied to roget's thesaurus. In:
  Wolff, K.E., Palchunov, D.E., Zagoruiko, N.G., Andelfinger, U. (eds.)
  Knowledge Processing and Data Analysis. pp. 150--163. Springer Berlin
  Heidelberg, Berlin, Heidelberg (2011)

\bibitem{Roth:2006}
Roth, C., Obiedkov, S.A., Kourie, D.G.: Towards concise representation for
  taxonomies of epistemic communities. In: Yahia, S.B., Nguifo, E.M.,
  Belohl{\'{a}}vek, R. (eds.) Concept Lattices and Their Applications, Fourth
  International Conference, {CLA} 2006, Tunis, Tunisia, October 30 - November
  1, 2006, Selected Papers. LNCS, vol.~4923, pp. 240--255. Springer (2006)

\bibitem{Rudin:2019}
Rudin, C.: Stop explaining black box machine learning models for high stakes
  decisions and use interpretable models instead. Nature Machine Intelligence
  \textbf{1}(5),  206--215 (2019)

\bibitem{Rudolph:2007}
Rudolph, S.: Using {FCA} for encoding closure operators into neural networks.
  In: Priss, U., Polovina, S., Hill, R. (eds.) Conceptual Structures: Knowledge
  Architectures for Smart Applications, 15th International Conference on
  Conceptual Structures, {ICCS} 2007, Sheffield, UK, July 22-27, 2007,
  Proceedings. Lecture Notes in Computer Science, vol.~4604, pp. 321--332.
  Springer (2007)

\bibitem{Shapley:1953}
Shapley, L.S.: A value for n-person games. Contributions to the Theory of Games
   \textbf{2}(28),  307--317 (1953)

\bibitem{Shrikumar:2017}
Shrikumar, A., Greenside, P., Kundaje, A.: Learning important features through
  propagating activation differences. In: Precup, D., Teh, Y.W. (eds.)
  Proceedings of the 34th International Conference on Machine Learning.
  Proceedings of Machine Learning Research, vol.~70, pp. 3145--3153. PMLR,
  International Convention Centre, Sydney, Australia (06--11 Aug 2017)

\bibitem{Srikant:95}
Srikant, R., Agrawal, R.: Mining generalized association rules. In: Dayal, U.,
  Gray, P.M.D., Nishio, S. (eds.) VLDB'95, Proceedings of 21th International
  Conference on Very Large Data Bases, September 11-15, 1995, Zurich,
  Switzerland. pp. 407--419. Morgan Kaufmann (1995)

\bibitem{Srikant:97}
Srikant, R., Agrawal, R.: Mining generalized association rules. Future Gener.
  Comput. Syst.  \textbf{13}(2-3),  161--180 (1997)

\bibitem{Strumbelj:2014}
Strumbelj, E., Kononenko, I.: Explaining prediction models and individual
  predictions with feature contributions. Knowl. Inf. Syst.  \textbf{41}(3),
  647--665 (2014)

\bibitem{Stumme:2001}
Stumme, G., Taouil, R., Bastide, Y., Lakhal, L.: Conceptual clustering with
  iceberg concept lattices. Proc. of GI-Fachgruppentreffen Maschinelles Lernen
  \textbf{1} (2001)

\bibitem{Tatti:2011}
Tatti, N., Moerchen, F.: Finding robust itemsets under subsampling. In: {ICDM}
  2011. pp. 705--714 (2011)

\bibitem{Valtchev:2000}
Valtchev, P., Missaoui, R.: {Similarity-based Clustering versus Galois lattice
  building: Strengths and Weaknesses}. In: Huchard, M., Godin, R., Napoli, A.
  (eds.) Contributions of the ECOOP'00 Workshop, ``Objects and Classification:
  a Natural Convergence'', European Conference on Object-Oriented Programming
  (2000). vol. Research Report LIRMM n.00095, p.~w13 (2000)

\end{thebibliography}

\end{document}